\documentclass[10pt,journal,compsoc]{IEEEtran}
\ifCLASSOPTIONcompsoc
  \usepackage[nocompress]{cite}
\else
  \usepackage{cite}
\fi
\ifCLASSINFOpdf
\else
\fi
\usepackage{amsmath}
\usepackage{amssymb}
\usepackage{bm}
\usepackage{graphicx}
\usepackage{subcaption}
\usepackage{multirow}
\usepackage{booktabs}

\usepackage{color}
\usepackage{xcolor}
\definecolor{citecolor}{RGB}{34,139,34}
\usepackage[pagebackref=true,breaklinks=true,letterpaper=true,colorlinks,citecolor=citecolor,bookmarks=false]{hyperref}

\newcommand{\vs}{\emph{vs.}}
\newcommand{\ie}{\emph{i.e., }}
\newcommand{\eg}{\emph{e.g., }}

\begin{document}
%
\title{The Equalization Losses: Gradient-Driven Training for Long-tailed Object Recognition}
%
%
%
%

\author{Jingru~Tan$^{\dagger}$,
        Bo~Li$^{\dagger}$,
        Xin~Lu, 
        Yongqiang~Yao$^*$,
        ~Fengwei~Yu,
        Tong He,\\
        and~Wanli~Ouyang~\IEEEmembership{Senior Member,~IEEE,}
\IEEEcompsocitemizethanks{\IEEEcompsocthanksitem Jingru Tan and Bo Li are with the Tongji University, Shanghai, China. \protect \\
E-mail: \{tjr120,1911030\}@tongji.edu.cn

\IEEEcompsocthanksitem Xin Lu, Fengwei Yu and Yongqiang Yao are with Sensetime Research, Shanghai, China. \protect \\
E-mail: \{luxin,yufengwei\}@sensetime.com, soundbupt@gmail.com

\IEEEcompsocthanksitem Wanli Ouyang is with University of Sydney and Shanghai AI Laboratory. \protect \\ Email: wanli.ouyang@sydney.edu.au

\IEEEcompsocthanksitem Tonghe is with Shanghai AI Laboratory. Email: tonghe90@gmail.com

\IEEEcompsocthanksitem $^*$ Corresponding author. $^{\dagger}$ Equal Contribution. }

}

\IEEEtitleabstractindextext{%
\begin{abstract}

   Long-tail distribution is widely spread in real-world applications.
   Due to the extremely small ratio of instances, tail categories often show inferior accuracy. 
   In this paper, we find such performance bottleneck is mainly caused by the imbalanced gradients, which can be categorized into two parts: (1) positive part, deriving from the samples of the same category, and (2) negative part, contributed by other categories. Based on comprehensive experiments, it is also observed that the gradient ratio of accumulated positives to negatives is a good indicator to measure how balanced a category is trained. Inspired by this, we come up with a gradient-driven training mechanism to tackle the long-tail problem: re-balancing the positive/negative gradients dynamically according to current accumulative gradients, with a unified goal of achieving balance gradient ratios. Taking advantage of the simple and flexible gradient mechanism, we introduce a new family of gradient-driven loss functions, namely equalization losses. 
   We conduct extensive experiments on a wide spectrum of visual tasks, including two-stage/single-stage long-tailed object detection (LVIS), long-tailed image classification (ImageNet-LT, Places-LT, iNaturalist), and long-tailed semantic segmentation (ADE20K). 
   Our method consistently outperforms the baseline models, demonstrating the effectiveness and  generalization ability of the proposed equalization losses.Codes will be released at \url{https://github.com/ModelTC/United-Perception}.
\end{abstract}




\begin{IEEEkeywords}
Long-tailed Object Recognition, Object Detection, Image Classification, Semantic Segmentation
\end{IEEEkeywords}}

\maketitle

\IEEEdisplaynontitleabstractindextext

%
\IEEEpeerreviewmaketitle

\IEEEraisesectionheading{\section{Introduction}\label{sec:introduction}}
\IEEEPARstart{O}{bject} recognition is one of the most fundamental tasks in computer vision. It is an important step in a host of visual challenges, including object detection, semantic segmentation, and object tracking. Despite this fact, the task remains an open problem, not least due to the discrepancy among the proportions of different categories. Current benchmarks such as ImageNet~\cite{imagenet2009deng}, PASCAL VOC~\cite{pascalvoc2010everingham}, COCO\cite{coco2014lin}, and Cityscapes~\cite{Cordts2016Cityscapes} are carefully collected with balanced annotations for each category, which contradicts the long-tailed Zipfian distribution in natural images. Although existing methods have achieved impressive results, we still can observe performance bottlenecks~\cite{oltr2019liu,lvis2019gupta,ade20k2017zhou} on various benchmarks, especially in the non-dominant classes with fewer samples. As substantiated by recent literature~\cite{eql2020tan,eqlv22021tan,seesawloss2021wang}, tail categories are easily overwhelmed by the head categories while learning on a dataset distributed off balance and diversely.



Previous approaches can be roughly categorised into two groups: data resampling~\cite{cas2016shen,wsl2018mahajan,lvis2019gupta} and cost-sensitive learning~\cite{ldam2019cao2019,class_balance_loss2019cui}. These methods address the above problems by either designing complex sampling strategies or adjusting loss weights. 
Although promising, most of these methods are designed based on categories' frequency and often suffer from several drawbacks: (1) those frequency-based methods are not robust enough due to widespread easy negative samples~\cite{focalloss2017lin} and redundant positive samples~\cite{class_balance_loss2019cui}. (2) the accuracy is sensitive to the predefined hyper-parameters.


\begin{figure*}[ht]
  \begin{center}
  \includegraphics[width=0.98\linewidth]{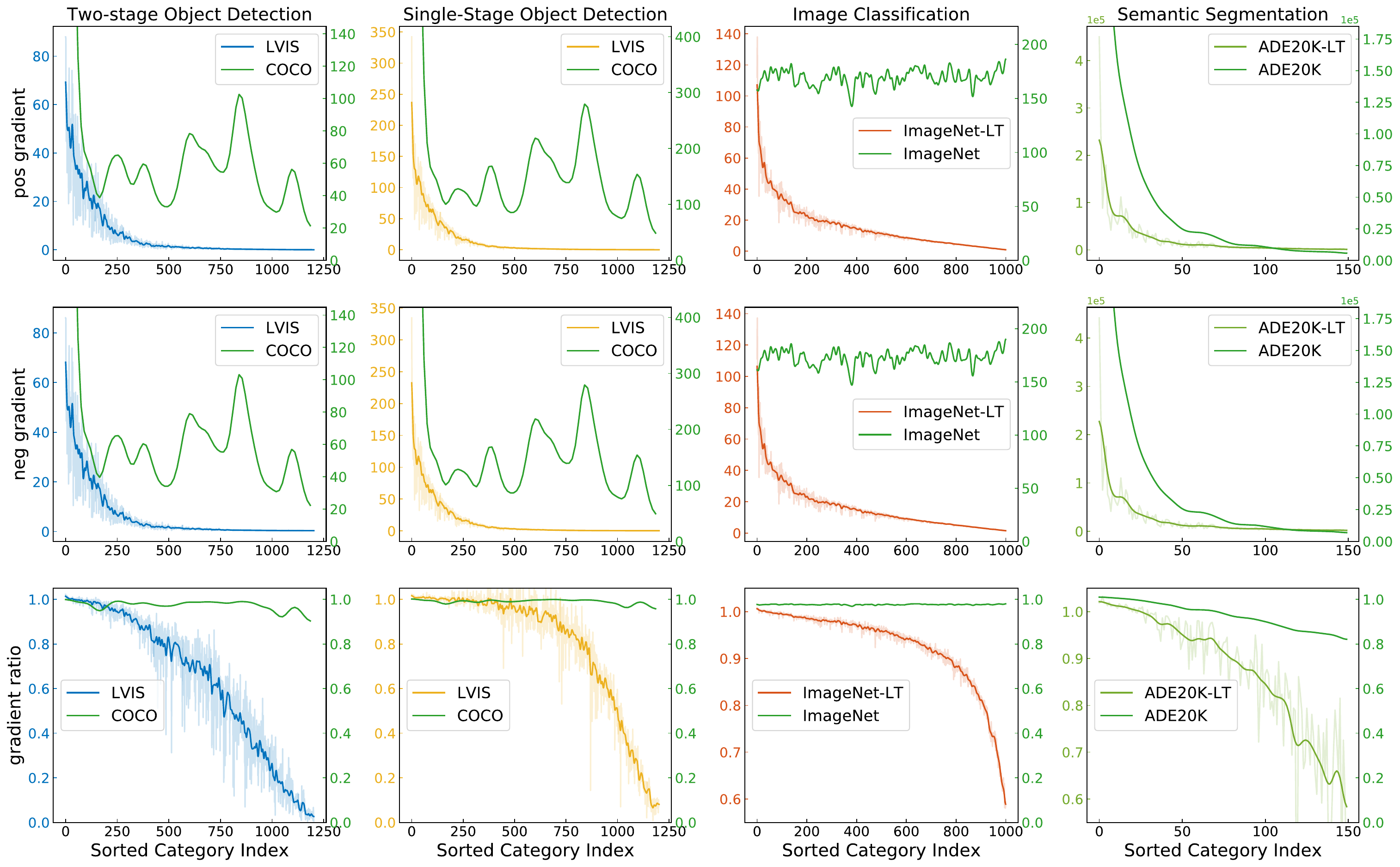}
  \end{center}
     \caption{Gradient observation on four different types of tasks. Each column is responsible for a specific task. We define the gradient of a sample as the derivative of the loss function with respect to its network output logits. For each category, we demonstrate the \textit{accumulated} gradient of positive samples (row 1), gradients of negative samples (row 2), and the gradient ratio of positive samples to negative samples (row 3). We sorted the category index according to its instance number. And we align COCO 80 categories with LVIS 1203 categories. The left and right y-axis are for imbalanced and balanced datasets.}
  \label{fig:grad_ob}
  \end{figure*}


%
In this paper, we tackle the problem of long-tailed recognition from a novel perspective. We start by analyzing the distribution of the accumulated gradients across different categories. Specifically, the gradients of one category consist of two parts: (1) the positive part, deriving from the samples of the same category, and (2) the negative part, contributed by other categories. Since the tail categories have limited positive samples, their positive gradients can be easily overwhelmed by the negative part. As illustrated in Fig.~\ref{fig:grad_ob} (bottom row), the gradient ratio of positive to negative examples are distributed off balance when trained on long-tailed datasets such as LVIS~\cite{lvis2019gupta}, ImageNet-LT~\cite{oltr2019liu}, and ADE20K-LT~\cite{ade20k2017zhou}. The gradients ratio is close to 1 for the head categories but is close to 0 for the tail categories. We hypothesize that such gradient imbalance in training is the main obstacle impeding tail classes from obtaining satisfactory performance.
%
%
Besides, we also conduct experiments on the well-balanced datasets such as COCO~\cite{coco2014lin}, ImageNet~\cite{imagenet2009deng}, and ADE20K~\cite{ade20k2017zhou}. It can be observed that all categories have a gradient ratio close to 1 without introducing any bias toward positives or negatives. Therefore, we believe that the gradient ratio can serve as a significant indicator of how balanced a category is trained.
Such a gradient-based indicator provides useful guidance to adjust gradients of the positive and negative parts, which can be easily plugged into different classifiers. To this end, we adapt it to various loss functions. (1) For binary cross-entropy loss (BCE), we introduce a \textbf{gradient-driven re-weighting} mechanism and propose the sigmoid equalization loss (Sigmoid-EQL). It treats the overall classification problem as a set of independent binary classification tasks.
%
Then the accumulative gradient ratio is used to up-weight the positive gradients and down-weight the negative gradients accordingly, aiming to balance the gradients of the two parts. 
(2) For cross-entropy loss (CE), we propose the softmax equalization loss (Softmax-EQL), which calibrates the decision boundary dynamically based on the statistics of the gradients. 
(3) For focal loss (FL)~\cite{focalloss2017lin}, we come up with the equalized focal loss (EFL) by decoupling the coefficients in \cite{focalloss2017lin} into category-agnostic and category-specific parts. By introducing the gradient into the category-specific parts, the model is able to focus more on the learning of rare categories.
Those losses do not rely on the pre-computed data statistics to determine the rebalancing terms. Instead, they control the training process in a dynamic way. This data-distribution agnostic property makes them more suitable for stream and realistic data. 

To demonstrate the effectiveness of our proposed method, comprehensive experiments have been conducted on various datasets and tasks. For object detection on the challenging LVIS~\cite{lvis2019gupta} benchmark, our proposed Sigmoid-EQL and Softmax-EQL outperform Mask R-CNN~\cite{maskrcnn2017he} by about 6.4\% and 5.7\% in terms of AP, respectively. Without introducing extra computation overhead, our approach improves the performance substantially. With the help of equalization losses, we won the first place both in COCO-LVIS challenge 2019 and 2020~\cite{tan20201st}. We also validate the effectiveness of our proposed EFL on the task of single-state object detection. Our method achieves 29.2\% AP, delivering significant improvements over state-of-the-art results. In addition to the effectiveness, the equalization losses also show strong generalization ability when transferring to other datasets and visual tasks. For example, equalization losses maintain huge improvements when moving from LVIS to Openimages~\cite{openimages2018kuznetsova} \textbf{without further hyper-parameters tuning}. In Openimages, Sigmoid-EQL and EFL outperform the baseline CE method by 9.1\% AP and 6.6\% AP, respectively. For the image classification task, Softmax-EQL achieves state-of-the-art results on three long-tailed image classification datasets (ImageNet-LT, Place-LT, and iNatrualist2018). 
We also evaluate our method on semantic segmentation using ADE20K~\cite{ade20k2017zhou}. Our proposed Sigmoid-EQL improves the powerful baseline, DeepLabV3+~\cite{deeplabv3plus2018chen}, by 1.56\% and 2.27\% in terms of mIoU and mAcc, respectively, showing strong generalization of gradient-driven losses to varying tasks. 

\section{Related Work}


\textbf{Long-tailed Object Recognition.} Common solutions for long-tailed image recognition are data re-sampling and loss re-weighting. Re-sampling methods under-sample the head categories~\cite{buda2018systematic, liu2008exploratory} or over-sample the tail categories~\cite{cas2016shen, wsl2018mahajan, bordersmote2005han, smote2002}. Re-weighting methods assign different weights to different categories~\cite{class_balance_loss2019cui, learning2017wang, learning2016huang, rangeloss2017zhang} or instances~\cite{focalloss2017lin, ghm2019li, meta_weight2019shu}. Decoupled training methods~\cite{crt2019kang, distalign2021zhang} address the classifier imbalance problem with a two-stage training pipeline by decoupling the learning of representation and classifier. In addition, margin calibration \cite{ldam2019cao2019, BalMS2020ren, seesawloss2021wang, logit_adjustment2021Aditya} inject category-specific margins into the CE loss to re-balance the logits distribution of categories. Recently, other works address the long-tailed problem from different perspectives such as transfer learning~\cite{oltr2019liu, chu2020feature, yin2019feature}, supervised contrastive learning~\cite{kang2020exploring, cui2021parametric, li2022targeted, zhu2022balanced}, ensemble learning~\cite{zhou2020bbn, wang2020long, cai2021ace, zhang2021test, li2022nested}, and so on. Some works adapt those ideas to object detection~\cite{lvis2019gupta}, including data re-sampling~\cite{lvis2019gupta, simcal2020wang}, loss re-weighting~\cite{eql2020tan, eqlv22021tan, wang2021adaptive}, decoupled training~\cite{simcal2020wang, bags2020li, distalign2021zhang}, margin calibration~\cite{seesawloss2021wang, pan2021model}, incremental learning~\cite{lst2020hu} and causal inference~\cite{tde2020tang}. Despite the efforts, most of them somehow utilize the sample number as the indicator of imbalance to design their algorithms. In contrast, we use the gradient as the indicator. It is more stable and precise thus reflects the models' training status better.

\vspace{0.2cm}
\noindent \textbf{Gradient as Indicator.} There are some works \cite{focalloss2017lin,ghm2019li}, that attempt to solve the imbalance problems from the gradient view. They use the \textit{instant} gradient to reflect the learning difficulty of a sample at a certain moment and determine its loss contribution dynamically, which can be viewed as an online version of hard negative example mining~\cite{shrivastava2016training}. Those methods are designed for the serious foreground-background imbalance problem.
Different from them, we use \textit{accumulative} gradients to reflect the imbalanced training status of categories. Our method is designed for long-tailed object recognition. Meanwhile, our method is complementary to theirs. We can solve the foreground-background imbalance problem and foreground-foreground (long-tailed) problem simultaneously by combining \textit{instant} gradients and \textit{accumulative} gradients indicators.



\section{Gradient Imbalance Problem}
\label{sec:gradient_imbalance_problem}
In this section, we introduce the imbalanced gradients, which we believe should be responsible for the inferior performance of the tail categories. It comes from the entanglement of instances and categories, which we will describe next. Building upon comprehensive experiments, we argue such gradient statistics can be served as an effective indicator to show the status of category classifiers. 

\noindent \textbf{Notation Definition.} Suppose we have a training set $\mathcal{X} = \{x_i, y_i\}_{N}$ with $C$ categories. Let the total instance number over the dataset be $N$ and $N = \sum_{j}^{C} n_j$, where $n_j$ is the instance number of category $j$. For each iteration, we have a batch of instances $\mathcal{I}$ with a batch size of $B$. $\mathcal{Y} \in \mathbb{R}^{B \times C}$ are the one-hot labels of the batch. We adopt a CNN $f$ with parameter $\theta$ as the feature extractor. Then the feature representations of the batch could be computed by $f(\mathcal{I}; \theta)$. A linear transformation is used as the classifier to output the logits: $\mathcal{Z} \in \mathbb{R}^{B \times C}$. $\mathcal{Z} = \bm{W}^T f(\mathcal{I}; \theta) + \bm{b}$, where $\bm{W}$ denotes the classifier weight matrix and $\bm{b}$ is the bias. 


We denote each image as an instance. $\bm{W}$ can be regarded as $C$ classifiers, each of which is responsible for bi-categorizing instances as one class. \textit{Each instance can be regarded as a positive sample for one specific category and a negative sample for the remaining $C-1$ categories.} We denote $y_{i}^{j} \in \{0, 1\}$ the label, which equals to 1 if the $i$-th instance belongs to the $j$-th category. 


\subsection{Entanglement of Instances and Categories}
\label{sec:entanglement}




The total number of positive samples $M_{j}^{pos}$ and negative samples $M_{j}^{neg}$ for the $j$-th classifier can be easily obtained: 

\begin{equation}
    M_{j}^{pos} = \sum_{i \in \mathcal{X}} y_{i}^{j}, \quad M_{j}^{neg} = \sum_{i \in \mathcal{X}} (1 - y_{i}^{j})
\label{eq:pos_neg_sample_number}
\end{equation}

 The ratio of the number of positive samples to the negative samples over the dataset is then:

\begin{equation}
    \frac{M_{j}^{pos}}{M_{j}^{neg}} \propto \frac{n_j}{N - n_j} \propto \frac{1}{\frac{N}{n_j} - 1} 
\label{eq:pos_neg_number_ratio}
\end{equation}



From Eq. \ref{eq:pos_neg_number_ratio}, we observe $M_{j}^{pos} \ll M_{j}^{neg}$ for the tail categories that have a very limited number of instances, indicating these categories often suffer from the extremely imbalanced ratio of positive to negative samples. Previous methods~\cite{class_balance_loss2019cui, ldam2019cao2019} address the problem by applying different loss weights or decision margins to different categories according to their sample numbers. However, they often fail to generalize well to other datasets because the sample numbers can not reflect the training status of each classifier well. For example, a large number of easy negatives samples and some redundant positive samples hardly contribute to the learning of the model. 
In contrast, we propose to use gradient statistics as our metric to indicate whether a category is in balanced training status. And we conjecture that there is a similar positive-negative imbalance problem in the gradient ratios of rare categories.

\subsection{Gradient Computation.}
We define the gradient over the batch $\mathcal{I}$ as the derivative of objective cost function $\mathcal{L}$ with respect to their logits $\mathcal{Z}$.  The gradient $\mathcal{G} = \frac{\partial \mathcal{L}}{\partial Z} \in \mathbb{R}^{B \times C}$ is corresponding to the gradients of all samples belonging to $C$ categories. We denote the gradient of a certain sample as $g_{i}^{j}$. Then the positive gradients $g(t)^{pos}_{j}$ and negative gradient $g(t)^{neg}_{j}$ of the category $j$ at iteration $t$ can be computed as follows:
\begin{equation}
    g(t)_{j}^{pos} = \sum_{i \in \mathcal{I}} y_{i}^{j} |g_{i}^{j}|, \quad g(t)_{j}^{neg} = \sum_{i \in \mathcal{I}} (1 - y_{i}^{j}) |g_{i}^{j}|
\label{eq:pos_neg_batch_gradient}
\end{equation}
The \textit{accumulated} positive gradients $G(T)^{pos}_{j}$ and negative gradients $G(T)^{neg}_{j}$ at iteration $T$ could be defined as:

\begin{equation}
     G(T)^{pos}_{j} = \sum_{t=0}^{T} g(t)_{j}^{pos}, \quad
     G(T)^{neg}_{j} = \sum_{t=0}^{T} g(t)_{j}^{neg}
 \end{equation}

For simplicity, we ignore $T$ in the notations and directly adopt $G^{pos}_{j}$ and $G^{neg}_{j}$ as the accumulated positive and negative gradients, respectively. The accumulated gradient ratio could be calculated by $G_j = \frac{G^{pos}_{j}}{G^{neg}_{j}}$. 


\subsection{Gradient Observation}
\label{sec:gradient_observation}

To validate our hypothesis that rare categories suffer from gradient imbalance problems, we collect gradient statistics during the training process across a wide spectrum of recognition tasks and datasets, including long-tailed image classification (ImageNet-LT~\cite{oltr2019liu}), two-stage/single-stage long-tailed object detection (LVIS~\cite{lvis2019gupta}), and long-tailed semantic segmentation (ADE20K~\cite{ade20k2017zhou}). The results are shown in Fig. \ref{fig:grad_ob}. 
We \textbf{consistently} observe four key phenomenons: (1) The \textit{positive} gradients $G^{pos}$ follow a long-tailed distribution. (2) The \textit{negative} gradients $G^{neg}$ follow a long-tailed distribution. (3) The gradient magnitudes of positive and negative over categories are different, leading their ratio $G$ also have the property of long-tailed distribution. (4) The gradient ratio $G$ of head categories is close to 1 while the ratio of tail categories is close to 0.

The x-axis of Fig. \ref{fig:grad_ob} is sorted by category instance numbers $n_j$. 
We notice that the positive gradients  have a positive correlation to the positive sample number, while the negative gradients do not have a positive correlation to the negative sample number. This is because tail categories get scarcely training by the model. Model hardly predicts those tail categories to be positives so most of their negative samples are easy samples. Although easy negative samples have small gradients, the effect accumulated from a large amount of them is not negligible. The observation of the gradient ratio proves that the category classifiers with fewer samples suffer from a more serious gradient imbalance  problem between positives and negatives, which validates our conjecture. For the head categories with abundant training samples, the received gradient ratio of the corresponding classifier is close to 1, indicating the classifier gives no inclination to positives or negatives, which we refer to as a balanced training status. For the tail classifier, the gradient ratio is close to 0, indicating the classifier is heavily biased towards negative, which we refer to as an imbalanced training status. It is worth noting that there is a vast number of background negative samples in single-stage object detection, as discussed in ~\cite{focalloss2017lin}. We find that the head categories still have a gradient ratio close to 1 in a well-trained model, as shown in Fig. \ref{fig:grad_ob} (column 3). This proves that the gradient indicator is more stable and reliable than the sample number.


As illustrated in Fig.~\ref{fig:grad_ob}, more experiments are conducted on several datasets that have close number of instances among different categories, including ImageNet~\cite{imagenet2009deng} and COCO~\cite{coco2014lin}. We observe the gradients are balanced distributed and the category classifiers have a balanced gradient ratio closing to 1. 

By observing the gradient statistics under imbalanced and balanced data distribution, we conclude that the gradient (\ie $G^{pos}_{j}$, $G^{neg}_{j}$) and the gradient ratio (\ie $G_j$) could serve as an important indicator, showing the training status of categories.

\section{The Equalization Losses}
Our central idea is to utilize gradient statistics as an indicator to reflect the training status of category classifiers and then adjust their training process dynamically. In this section, by applying this idea to several loss functions, such as binary cross-entropy loss, cross-entropy loss, and focal loss, we introduce a new family of gradient-driven loss functions, namely the equalization losses. 

\subsection{Sigmoid Equalization Loss}

\subsubsection{Binary Cross-Entropy}
Binary cross-entropy (BCE) loss estimates the probability of each category \textit{independently} using $C$ sigmoid loss functions. Specifically, in a batch of instances $\mathcal{I}$, the classifier outputs the estimated probability $\mathcal{P} \in \mathbb{R}^{B \times C}$ by applying a sigmoid activation function to the logits $\mathcal{Z}$ and $\mathcal{P} = \sigma(\mathcal{Z})$. We define the estimated probability of a certain sample as $p_{i}^{j} \in [0,1]$. For this sample, the loss term is computed as \footnote{For neatness, we ignore the superscripts and subscripts of $y$ and $p$. Unless otherwise stated, they are also be ignored in the formula of CE and focal loss for a certain sample.}:



\begin{equation}
  \mathrm{BCE}(p, y) =
  \begin{cases}
      -\mathrm{log}(p) & \text{if } y = 1 \\
      -\mathrm{log}(1-p)  & \text{otherwise}
  \end{cases} 
\end{equation}

\noindent Follow the notation in \cite{focalloss2017lin}, we define $p_\mathrm{t}$ as:

\begin{equation}
  p_\mathrm{t} =
  \begin{cases}
      p & \text{if } y = 1 \\
      1 - p  & \text{otherwise}
  \end{cases} 
\end{equation}

\noindent then we can rewrite $\mathrm{BCE}(p, y) = \mathrm{BCE}(p_\mathrm{t}) = -\mathrm{log}(p_\mathrm{t})$. And the final loss contribution could be calculated by summing up the loss values from all samples:
\begin{equation}
  \mathrm{L}(\mathcal{P}, \mathcal{Y}) = \sum_{i \in \mathcal{I}} \sum_{j = 1}^{C} \mathrm{BCE}(p_\mathrm{t})
\end{equation}

The probability of each category in the BCE is estimated independently without cross normalization. This property makes the binary cross-entropy suitable for tasks that consist of a set of independent sub-tasks, such as object detection, and multi-label image classification.

\subsubsection{Gradient-Driven Re-weighting}
Under long-tailed distribution, models are in an unbalanced training status. As mentioned in Section \ref{sec:gradient_observation}, the \textit{accumulated} gradient ratio $G_j$ can reflect the training status of that category. Therefore we adopt it to adjust the training process for each sub-task in BCE independently and equally. Concretely, we propose a gradient-driven re-weighting mechanism in which we up-weight the positive gradients and down-weight negative gradients for each classifier dynamically.  This re-weighting strategy aims to make the gradient ratio as close to 1 as possible.   

We denote $q_j$ as the weight term for positive samples of category $j$ and $r_j$ for negative samples. We propose the formulation as:

 \begin{equation}
    r_j = f(G_j) 
 \end{equation}

 \begin{equation}
    q_j = 1 + \alpha (1 - r_j) 
 \end{equation}

 \noindent where $f(\mathord{\cdot})$ is a mapping function to remap the value of gradient ratio to a more controllable range. Basically, for a small gradient ratio with imbalanced training status, we will have a small $r_j$ for negative gradients and a big $q_j$ for positive gradients to re-balance the training status of the current category. Simple mapping functions\footnote{We treat the gradient ratio with $g = 1$ as the most balanced status. So we also clip the value to [0, 1] after mapping for simplicity and think the classifier with $g > 1$ is also balanced.} like linear mapping: $f(x) = x$, exponential mapping: $f(x) = x^2$ or square root mapping: $f(x) = \sqrt{x}$ could be chosen. More sophistic mapping functions are also feasible, like the sigmoid-like mapping function: $f(x) = \frac{1}{1 + e^{-\gamma(x - \mu)}}$.

We name this novel gradient re-weighting mechanism Sigmoid Equalization Loss (Sigmoid-EQL), and the formula of the loss is:
\begin{equation}
  \mathrm{L}(\mathcal{P}, \mathcal{Y}) = \sum_{i \in \mathcal{I}} \sum_{j = 1}^{C} (q_{j} y_{i}^{j} + r_{j} (1-y_{i}^{j})) \mathrm{BCE}(p_\mathrm{t})
\end{equation}

\subsection{Softmax Equalization Loss}
\subsubsection{Cross-Entropy}
Cross-entropy (CE) loss is also a widely-used loss function in recognition tasks. The loss term of CE for a sample is:
\begin{equation}
  \mathrm{CE}(p, y) = -y\mathrm{log}(p)
  \label{eq:ce_formula}
\end{equation}

Different from BCE, CE uses a \textit{cross normalization} operation $\mathrm{softmax}$ to derive the estimated probability:
\begin{equation}
   p_{i}^{j} = \frac{e^{z_{i}^{j}}}{\sum_{k=1}^{C} e^{z_{i}^{k}}} 
\end{equation}

It is worth noting that although the CE only calculates the loss of the positive sample for an instance, the gradient will flow back to logits of negative samples because of the softmax function. This operation introduces explicit competition between categories. So it is useful for tasks that require a single output category, such as image classification and semantic segmentation. During inference, for instance $i$, we choose its predicted output category $y_{i}^{\prime}$ by $\mathrm{argmax}$ operation:

\begin{equation}
  y_{i}^{\prime} = \mathrm{argmax}_j p_{i}^{j}
\end{equation}







\subsubsection{Gradient-based Margin Calibration}
In the situation that the tasks require a single output category, we can no longer treat category classifiers as independent sub-tasks as BCE does. The accurate rank between categories is crucial so the learning of all samples in an instance should be adjusted jointly. Under the long-tailed situation, the CE loss prefers to predict instances as head categories because they received more gradients. The optimization direction is dominated by the head categories thus they occupy a broad area in the decision space.

As discussed in \cite{logit_adjustment2021Aditya}, the neural network models the output $p$ as a posterior probability $p(y|x)$ of $y$ given image $x$. According to Bayes' theorem:

\begin{equation}
  \label{eq:decision_func}
  p(y|x) \propto p(y)p(x|y)
\end{equation}

\noindent
$p(y)$ is called prior probability. For category $j$, its prior probability is defined as $p(y_j)$ which is proportional to the sample number $n_j$ and the accumulative positive gradients $G_{j}^{pos}$:

\begin{equation}
  p(y_j) \propto n_j \propto G_{j}^{pos}
\end{equation}

Obviously, head categories have a higher prior probability, so the model is inclined to predict head categories according to the decision function Eq. \ref{eq:decision_func}. This prediction preference harms the performance of tail categories.

To alleviate the influence of the prior probability, we propose gradient-based margin calibration, a strategy that injects accumulated gradient to conventional softmax so that model is able to be aware of observed data distribution. As a result, the model can output a fair prediction result by adopting the dynamical calibration strategy.

The probability of a sample in our gradient-based margin calibration method could be calculated by:

\begin{equation}
   p_{i}^{j} = \frac{(G_j^{pos})^{\pi} e^{z_{i}^{j}}}{\sum_{k=1}^{C} (G_k^{pos})^{\pi}e^{z_{i}^{k}}} 
\end{equation}

\noindent where $\pi$ is a hyper-parameter that controls the calibration degree. It is worth noting that we apply the accumulative positive gradient $G_{j}^{pos}$ as the gradient indicator because it could indicate the prior probability better than the gradient ratio $G_{j}$. By replacing the conventional softmax function with our gradient-based margin calibration method, we propose the Softmax Equalization Loss (Softmax-EQL). Softmax-EQL shares a similar idea with Balance Meta-Softmax Loss~\cite{BalMS2020ren}, Logit Adjustment~\cite{logit_adjustment2021Aditya} Loss, and Seesaw Loss~\cite{seesawloss2021wang}, but they use the sample number as the indicator. We will prove that the gradient indicator reflects the relation between categories more precisely and achieve better results.

\subsection{Equalized Focal Loss}

In this section, we will demonstrate that our gradient indicator can be applied not only to the simple BCE and CE losses but also to more complicated losses (\eg focal loss~\cite{focalloss2017lin} and quality focal loss~\cite{li2020generalized})  in more difficult tasks (single-stage long-tailed object detection). Single-stage detection has a simple and fast pipeline that is very prevalent in real-world applications. In contrast to two-stage methods that incorporate a region proposal network (RPN \cite{faster_rcnn2015ren}) to filter out most background samples before feeding proposals to the final classifier, single-stage detectors directly detect objects over a regular, dense set of candidate locations. Due to this dense prediction schema, the extreme foreground-background imbalance is introduced. This means that, under the long-tailed distribution, the single-stage detectors have to solve the imbalance problem between foreground-background instances and between foreground categories' instances simultaneously. 
 
\subsubsection{Focal Loss}
\label{sec:revisiting_fl}


Focal loss \cite{focalloss2017lin} is a conventional solution to the
foreground-background imbalance problem. Similar to the definition of BCE and CE, the formula of focal loss for a certain sample is:
\begin{equation}
  \mathrm{FL}\left(p_{\mathrm{t}}\right)=-\alpha_{\mathrm{t}}\left(1-p_{\mathrm{t}}\right)^{\gamma} \log \left(p_{\mathrm{t}}\right)
  \label{eq:focal_loss}
\end{equation}


As declared in \cite{focalloss2017lin}, $\alpha_{\mathrm{t}}$ is a balance factor that balances the importance of positive samples and negative samples. The modulating factor $\left(1-p_{\mathrm{t}}\right)^{\gamma}$ is the key component of focal loss. It down-weights the loss of easy samples and focuses on the learning of hard samples by the predicted $p_{\mathrm{t}}$ and the focusing parameter $\gamma$. Since most samples of background instances are easy to classify, focal loss greatly weakens the influence of these background instances, thus focusing more on the learning of foreground instances. It could be concluded from Eq. \ref{eq:focal_loss} that the larger the $\gamma$, the more focus on the learning of hard positive samples, which is suitable for more serious positive-negative imbalance problems.



\subsubsection{Gradient Driven Modulating}
\label{sec:efl_definition}



However, focal loss fails to solve the foreground categories imbalance problem under the long-tailed data distribution (see~Table~\ref{tab:main_results}). In long-tailed dataset (\ie LVIS), as described in Section \ref{sec:gradient_observation}, different categories suffer from different degrees of positive-negative gradient imbalance problems. Rare categories have more severe imbalance problems than frequent ones. Therefore, focal loss with the constant modulating factor may be not appropriate for all these imbalance problems. Based on these analysis, we propose Equalized Focal Loss (EFL) by designing a gradient-driven modulating mechanism for the focal loss. Concretely, we introduce two category-relevant factors, \ie the focusing factor and the weighting factor. Those two factors are controlled by our gradient indicator to dynamically handle positive-negative imbalance problems of different degrees.

\vspace{2mm} \noindent
\textbf{Focusing Factor.} We assign large focusing factors to rare categories to alleviate their serious positive-negative gradient imbalance problems. And for frequent categories with slight imbalance problem, a small focusing factor is proper. We formulate the loss of a sample belonging to the $j$-th category as:
\begin{equation}
  \mathrm{EFL}\left(p_{\mathrm{t}}\right)=-\alpha_{\mathrm{t}}\left(1-p_{\mathrm{t}}\right)^{\gamma^j} \log \left(p_{\mathrm{t}}\right)
  \label{eq:efl_loss}
\end{equation}

Specifically, we decouple the gradient-driven focusing factor $\gamma^j$ into two components which are a categories-agnostic parameter $\gamma_b$ and a categories-specific parameter $\gamma_v^j$:
\begin{equation}
  \begin{aligned}
  \gamma^j &=\gamma_b + \gamma_v^j \\
  &=\gamma_b + s\left(1 - G_j\right)
  \end{aligned}
  \label{eq:gamma_formula}
\end{equation}
where $\gamma_b$ controls the basic behavior of the loss.
And the parameter $\gamma_v^j \geq 0$ is a variable associated with the imbalance degree of the $j$-th category. We reflect the imbalance degree (training status) of the $j$-th category by the accumulated gradient ratio $G_j$. And we directly apply the linear mapping function $1 - G_j$ to make the focusing factor keep a negative correlation to the imbalance degree. The hyper-parameter $s$ is a scaling factor that determines the upper limit of $\gamma^j$ in EFL.
Compared with focal loss, EFL could handle the positive-negative imbalance problem of each category independently and dynamically.




\vspace{2mm}
\noindent
\textbf{Weighting Factor.} Even with the focusing factor $\gamma^j$, there is still an obstacle degrading the performance: For a binary classification task, a larger $\gamma^j$ is suitable for a more severe positive-negative imbalance problem. However, in the multi-class case (see Eq. \ref{eq:efl_loss}), for two samples of different classes that have the same logits, the larger the value of $\gamma^j$, the smaller the loss. It leads to the fact that when we want to increase the concentration on learning a category, we have to sacrifice part of its loss contribution in the overall training process. Such a dilemma prevents rare categories from achieving excellent performance. Basically, we expect rare hard samples to make more loss contributions than frequent hard ones.

We propose the weighting factor to alleviate the problem by re-weighting the loss contribution of different categories. Similar to the focusing factor, indicated by the gradient, the weighting factor is assigned a large value for rare categories to raise their loss contributions while keeping close to 1 for frequent categories.
Specifically, we set the weighting factor of the $j$-th category as $w^{j}$ and the final formula of EFL is:

\begin{equation}
  \mathrm{EFL}\left(p_{\mathrm{t}}\right)=-w^{j}\left(1-p_{\mathrm{t}}\right)^{\gamma^{j}} \log \left(p_{\mathrm{t}}\right)
  \label{eq:total_efl_loss}
\end{equation}

\noindent where $w^{j}$ shares the same variable $\gamma_b$ and $\gamma_v^j$ as the focusing factor:
\begin{equation}
  w^{j}=\frac{\gamma_b + \gamma_v^j}{\gamma_b}
\end{equation}

The focusing factor and the weighting factor make up the categories-relevant modulating factor in EFL. It enables the classifier to dynamically adjust the loss contribution of a sample based on our gradient indicator.
We will show later in experiments that both the focusing factor and the weighting factor play significant roles in EFL.
Meanwhile, in the balanced data distribution, all categories in EFL will have balanced training status with all $G$ close to 1. Then the $\gamma^j=\gamma_b$ makes the EFL equivalent to focal loss.
Such an appealing property makes EFL could be applied well with various data distributions. 

What's more, the gradient-driven modulating factor is also applicable in other focal loss series such as the quality focal loss (QFL) \cite{li2020generalized}. The novel loss is denoted as the equalized quality focal loss (EQFL), and the formulated of a sample in EQFL of the $j$-th category is:
\begin{equation}
  \mathrm{EQFL}\left(p\right)=-m^j\left(y^{\prime}\log\left(p\right) + \left(1-y^{\prime}\right)\log\left(1-p\right)\right)
  \label{eq:eqfl_loss}
\end{equation}
\noindent where $m^j$ is the specific form of the gradient-driven modulating factor in EQFL:
\begin{equation}
  m^j=w^j\left(\left|y^{\prime}-p\right|\right)^{\gamma^j}
\end{equation}

The weighting factor $w^j$ and the focusing factor $\gamma^j$ are the same as in EFL. It should be noticed that $y{\prime} \in [0, 1]$ is the IoU score for a positive sample and 0 for a negative sample as declared in \cite{li2020generalized}. As we will show in Section \ref{sec:single_stage_analysis}, combined with the advanced YOLOX \cite{ge2021yolox} detectors, our proposed EQFL achieves impressive performance improvement in solving the long-tailed problem.

\begin{table*}
  \centering
  \setlength\tabcolsep{16pt}
  \begin{tabular}{c |l |c c c c c }
    \toprule
     framework & method & AP & AP$_r$ & AP$_c$ & AP$_f$ & AP$_b$ \\
    \midrule
    \multirow{7}{*}{R50 Mask R-CNN} & CE baseline & 19.2 & 0 & 17.2 & 29.5 & 20.0 \\
     & Grad-Blocking~\cite{eql2020tan} & 21.6 & 3.8 & 21.7 & 29.2 & 22.5 \\
    &RFS~\cite{lvis2019gupta} & 24.4 & 14.6 & 23.7 & 29.4 & 25.0 \\
    &BAGS~\cite{bags2020li} & 25.1 & 14.7 &\textbf{24.9} & 29.8 & 25.5 \\
    &Seesaw Loss~\cite{seesawloss2021wang} & 24.7 & 14.8 & 24.0 & 29.8 & 25.3 \\
    \cmidrule(l){2-7}
    & Softmax-EQL (Ours) & 25.3 & 17.4 & 24.5 & 29.7 & 25.5 \\
    & Sigmoid-EQL (Ours) & \textbf{25.5} & \textbf{17.7} & 24.3 & \textbf{30.2} & \textbf{26.1} \\
    \midrule
    \multirow{7}{*}{R101 Mask R-CNN} & CE baseline & 20.8 & 1.4 & 19.4 & 30.9 & 21.7 \\
    & Grad-Blocking~\cite{eql2020tan} & 22.9 & 3.7 & 23.6 & 30.7 & 24.2 \\
    &RFS~\cite{lvis2019gupta} & 25.8 & 17.4 & 24.7 & 30.7 & 26.8 \\
    &BAGS~\cite{bags2020li} & 26.4 & 17.2 & 26.0 & 31.0 & 26.9 \\
    &Seesaw Loss~\cite{seesawloss2021wang} & 26.4 & 16.7 & 25.7 & 31.4 & 27.0 \\
    \cmidrule(l){2-7}
    &Softmax-EQL (Ours) & 26.5 & 18.0 & 25.7 & 31.0 & 26.9 \\
    &Sigmoid-EQL(Ours) & \textbf{27.2} & \textbf{20.6} & \textbf{25.9} & \textbf{31.4} & \textbf{27.9} \\
  \bottomrule
   \end{tabular}
   \caption{Comparison with state-of-the-art methods on \textbf{LVIS v1.0} \texttt{val} set. All models are trained in an end-to-end way with random sampler by a 2x schedule. AP and AP\textsubscript{\textit{b}} denotes mask AP and box AP respectively.}
   \label{tab:comparison_sota_lvis1}
\end{table*}


\section{Experimental Results}
In this section, we conduct extensive experiments on the equalization losses. To show the versatility of the gradient indicator, we present experiments on four popular recognition tasks: two-stage long-tailed object detection, single-stage long-tailed object detection, long-tailed image classification, and long-tailed semantic segmentation. For a specific vision task, we first introduce the dataset setup and the evaluation metric. Then we describe the important implementation details. Finally, we 
present the main results and analysis. 


\subsection{Two-Stage Long-tailed Object Detection}

Object detection requires detecting all possible objects with a set of pre-defined categories. The evaluation metric (AP) is evaluated class-wisely. Therefore, we can model the detection task as a set of independent sub-tasks, each for one specific category, whose probability is estimated by a sigmoid loss function. In large-scale object detection, the concepts of categories are not always mutually exclusive but show the federated characteristic. Compared to the softmax loss function, the sigmoid loss function gives more flexible control of the relationship between categories. Thus, in this object detection task, we mainly study the Sigmoid-EQL. 

\subsubsection{Datasets and Evaluation Metric}
LVIS~\cite{lvis2019gupta} is a challenging benchmark for long-tailed object recognition. It provides precise bounding box and mask annotations for various categories with long-tailed distribution. We mainly perform experiments on the recently released challenging LVIS v1.0 dataset. It consists of 1203 categories. We train our models on the \texttt{train} set, which contains about 100k images with 1.3M instances. In addition to widely-used metric AP across IoU threshold from 0.5 to 0.95, LVIS also reports AP\textsubscript{r} (rare categories with 1-10 images), AP\textsubscript{c} (common categories with 11-100 images), AP\textsubscript{f} (frequent categories with $>$ 100 images). Since LVIS is a federated dataset, categories are not annotated exhaustively. Each image have two more types of labels: \texttt{pos\_category\_ids} and \texttt{neg\_category\_ids}, indicating which categories are or are not present in that image. Detection results that do not belong to those categories will be ignored for that image. We report results on the \texttt{val} set of 20k images. 

\subsubsection{Implementation Details}
Models are trained using SGD with a momentum of 0.9 and a weight decay of 0.0001. The ResNet~\cite{resnet2016he} backbones are initialized by ImageNet pre-trained models. Following the convention, scale jitter and horizontal flipping are used in training and no test time augmentation is used. We use a total batch size of 16 on 16 GPUs (1 image per GPU) and set the initial learning rate to 0.02. According to ~\cite{lvis2019gupta}, the maximum number of detection per image is up to 300, and the minimum score threshold is reduced to 0.0001 from 0.01. For Sigmoid-EQL, we initialize the bias of the last fully-connected classification layer (fc-cls) with values of 0.001 to stabilize the training at the beginning. We also add a branch for detecting objectiveness instead of the concrete category to reduce false positives. In the training phase, this branch treats all other categories' positive samples as its positive samples. In the inference phase, the estimated probability of other categories become: $p^{\prime}_{j} = p_{j} * p_{obj} $. Where $p_{obj}$ is the probability of a proposal being an object. Note that, the proposed gradient-driven mechanisms (re-weighting or calibration) are not applied in the objectiveness branch. 

\subsubsection{Main Results}

We compare our method with several leading methods on LVIS v1.0 \texttt{val}. All models are trained in an end-to-end way with random sampler by a 2x schedule. The results are shown in Table \ref{tab:comparison_sota_lvis1}. On the R50 Mask R-CNN framework, the Sigmoid-EQL outperforms the CE loss and Grad-Blocking~\footnote{To avoid name conflict, we rename the original equalization loss~\cite{eql2020tan} to Grad-Blocking.}~\cite{eql2020tan} by a large margin, 6.3\% AP and 3.9\% AP respectively. Note that RFS~\cite{lvis2019gupta} repeats images that contain tail categories in each epoch, so it increases the total training time. Instead, our method only uses a random sampler and does not increase the training time, and achieves better results, 25.5\% AP \vs~24.4\% AP. Compared with the sample-number-based methods BAGS~\cite{bags2020li} and Seesaw Loss~\cite{seesawloss2021wang}, Sigmoid-EQL outperforms them with a non-trivial improvement, 0.4\% AP and 0.8\% AP respectively. We also provide the result of Softmax-EQL and compare it with Seesaw Loss. Without the hard logit mining, \ie compensation factor like in Seesaw Loss, our Softmax-EQL achieves better performance than it, demonstrating that the accumulated gradient ratio is a more superior indicator than the accumulated sample number. Meanwhile, under a stronger backbone (\eg ResNet-101), our proposed method still achieves higher overall AP across different algorithms.

\begin{table}
  \centering
  \setlength\tabcolsep{6pt}
  \begin{tabular}{c c c | c c c c c}
        \toprule
        obj? & neg? & pos? & AP & AP\textsubscript{\textit{r}} & AP\textsubscript{\textit{c}} & AP\textsubscript{\textit{f}} & AP\textsubscript{\textit{b}} \\
        \midrule
        & & & 16.1 & 0 & 12.0 & 27.4 & 17.2 \\
        \checkmark & & & 18.1 & 1.9 & 16.4 & 28.3 & 19.0\\
        \checkmark & \checkmark & & 19.7 & 7.3 & 17.6 & 27.6 & 20.5 \\
        \checkmark & & \checkmark & 23.3 & 12.2 & 22.8 & \textbf{28.8} & 23.9 \\
        \checkmark & \checkmark & \checkmark & \textbf{23.7} & \textbf{14.9} &\textbf{22.8} & 28.6 & \textbf{24.2} \\
        \bottomrule 
  \end{tabular}
  \caption{Effect of different components. \textit{obj} for adding the objectiveness branch, \textit{neg} for reweighing negative gradients, \textit{pos} for reweighing positive gradients.}
  \label{tab:component_analysis}
\end{table}

\begin{table}
  \centering
  \setlength\tabcolsep{4pt}
  \begin{tabular}{c | c c c c | c c c c }
        \toprule
        \multirow{2}[3]{*}{function type}& \multicolumn{4}{|c|}{neg \checkmark} & \multicolumn{4}{c}{neg \checkmark pos \checkmark} \\
        \cmidrule(l){2-9} 
         & AP & AP\textsubscript{r} & AP\textsubscript{c} & AP\textsubscript{f} & AP & AP\textsubscript{r} & AP\textsubscript{c} & AP\textsubscript{f} \\
        \midrule
        sqrt ($y=\sqrt{x}$)& 18.4 & 2.1 & 17.6 & 28.2 & 21.4 & 6.2 & 21.0 & 28.6 \\
        linear ($y = x$)& 18.8 & 2.0 & 16.9 & \textbf{28.3}  & 22.6 & 10.0 & 22.0 & 28.7 \\
        exp ($y = x^2$)& 19.1 & 2.1 & 17.6 & 28.2  & 23.2 & 11.9 & 22.7 & \textbf{28.8} \\
        sigmoid-like & \textbf{19.7} & \textbf{7.3} & \textbf{17.6} & 27.6 & \textbf{23.7} & \textbf{14.9} & \textbf{22.8} & 28.6  \\
        \bottomrule
  \end{tabular}
  \caption{Comparison between different mapping functions}
  \label{tab:mapping_func}
\end{table}

\subsubsection{Analysis}

We conduct extensive experiments to give a thorough analysis of sigmoid-EQL. We use a 1x training schedule with a random sampler if not mentioned. 

\vspace{0.2cm}
\noindent \textbf{Component Effect.} The effect of each component in Sigmoid-EQL is shown in Table \ref{tab:component_analysis}. The baseline model performs poorly on the tail classes, resulting in 0\% AP and 12.0\% AP for rare and common categories. And the performance gaps between head and tail classes are very large. Adding an objectness branch helps all categories to some extent, improving the overall AP by 2.0\% but not very much for the rare categories since the main problem for them is the unbalanced positive and negative gradients, \ie. their positive gradients are overwhelmed by negative gradients cause by a vast number of negative samples. By down-weighting the influence of negative gradients, their accuracy is boosted significantly (5.4\% AP for rare categories).  Further up-weighting the positive gradient helps to achieve a more balanced ratio of positive to negative gradients. It brings a 7.6\% AP performance boost for rare categories. 
 With these three components, we achieve a 23.7\% AP, outperforming the baseline model 16.1\% AP by a large margin without any re-sampling techniques.
 It is worth noting that only up-weighting positive gradients already achieved a significant improvement but the AP of tail categories is limited by the massive discouraging negative gradients.
 These ablation experiments verified the effectiveness of the Sigmoid-EQL.

\vspace{0.2cm}
 \noindent \textbf{Mapping Functions.} In Table~\ref{tab:mapping_func}, we show the ablation study of the mapping function in Sigmoid-EQL. All those mapping functions improves the performances while sigmoid-like function $f(x) = \frac{1}{1 + e^{-\gamma(x - \mu)}}$ stands out from the others.  It has a value like an inflection point $\mu$, which we can treat as a high enough gradient ratio. As the gradient ratio increases, this function first changes very slowly when the ratio is far small than the inflection point. Then it increases dramatically when the ratio gets closed to the inflection point. And it slows down its changes again after becoming larger than the inflection point. The sigmoid-like mapping functions consistently achieve higher performance on both overall AP and AP$_r$. By default, we choose the inflection point $\mu$ as 0.8, and set $\gamma = 12$. 

 \vspace{0.2cm}
\noindent \textbf{Stability under Longer Training.} To verify the stability ability across different backbones and training schedules. We conduct experiments with larger models by a 3x schedule.  The results are present in Table \ref{tab:longer_schedule}. Note that training Mask R-CNN baseline (CE loss) with a longer schedule does not help rare categories a lot (Table \ref{tab:comparison_sota_lvis1} \vs Table \ref{tab:longer_schedule}), the AP of rare categories is still bad because rare categories are heavily suppressed by the negative gradients caused by the entanglement of instances and tasks. In contrast, with the proposed Sigmoid-EQL, the performance can be further improved from 25.5\% AP to 26.2\% AP on the R50 backbone. We do not observe the over-fitting of tail classes in such a long schedule. And when using a large ResNet-101 backbone, the gap between Mask R-CNN and Sigmoid-EQL still holds.

\begin{table}
  \centering
  \setlength\tabcolsep{6pt}
  \begin{tabular}{l |l | c c c c}
        \toprule
        backbone & method & AP & AP$_r$ & AP$_c$ & AP$_f$ \\
        \midrule
        \multirow{2}{*}{ResNet-50} & CE baseline & 20.5 & 2.0 & 19.0 & 30.3 \\
        & Sigmoid-EQL & \textbf{26.2} & \textbf{19.1} & \textbf{25.0} & \textbf{30.7} \\
        \midrule
        \multirow{2}{*}{ResNet-101} & CE baseline & 21.7 & 1.6 & 20.7 & 31.7 \\
        & Sigmoid-EQL & \textbf{27.5} & \textbf{20.5} & \textbf{26.2} & \textbf{32.0} \\
        \bottomrule
  \end{tabular}
  \caption{Results of larger backbones with a longer 3x schedule. A Random sampler is used. The models are trained with a total of 36 epochs, and the learning rate is divided by 10 at the 28th epoch and 34th epoch respectively.}
  \label{tab:longer_schedule}
\end{table}

\begin{table}
  \centering
  \setlength\tabcolsep{6pt}
  \begin{tabular}{c | l | c c c c c}
    \toprule
     strategy & method & AP & AP$_r$ &
     AP$_c$ & AP$_f$ & AP$_b$\\
     \midrule
     \multirow{2}{*}{Baseline} & CE & 16.1 & 0.0 & 12.0 & 27.4 & 16.7 \\
      & BCE & 16.5 & 0.0 & 13.1 & 27.3 & 17.2 \\
     \midrule
     \multirow{3}{*}{Decoupled} & LWS~\cite{crt2019kang} & 17.0 & 2.0  & 13.5 & 27.4 & 17.5 \\
      &  cRT~\cite{crt2019kang} & 22.1 & 11.9 & 20.2 & \textbf{29.0} & 22.2 \\
      & BAGS~\cite{bags2020li} & 23.1 & 13.1 & 22.5 & 28.2 & 23.7 \\
     \midrule
     \multirow{2}{*}{End-to-End} & Softmax-EQL & 23.1 & \textbf{15.3} & 22.1 & 27.6 & 23.3 \\
      & Sigmoid-EQL & \textbf{23.7} & 14.9 &\textbf{22.8} & 28.6 & \textbf{24.2} \\
     \bottomrule
  \end{tabular}
     \caption{
     Comparison with decoupled training methods. For cRT and LWS, they use the class-balance sampler to fine-tune their model in the second stage, and BAGS uses a random sampler following the original paper.}
  \label{tab:compare_e2e_decouple}
\end{table}

\vspace{0.2cm}
\noindent \textbf{Comparison with Decoupled Training algorithms.} We mainly compare our method with three decoupled training methods (cRT~\cite{crt2019kang}, LWS~\cite{crt2019kang}, and BAGS~\cite{bags2020li}). The results are present in Table \ref{tab:compare_e2e_decouple}. The decoupled training models are first initialized from naive softmax baseline (CE loss), then re-train their classifier layer (fc-cls) for another 12 epoch with other layers frozen, resulting in a total 24 epoch training. Those decoupled training methods all improve the AP, mainly for tail classes. The improvements brought by LWS are limited. We conjecture it is because that LWS only learns a scaling factor to adjust the decision boundary of the classifier but the classifier itself is not good and imbalanced. Our Sigmoid-EQL achieves a overall 23.7\% AP, outperforms the LWS, cRT, and BAGS by 6.7\%, 1.6\% and 0.6\%, respectively. It is worth noting that Sigmoid-EQL does not require the extra fine-tuning stage, and the representation and classifier are learned jointly.

\begin{figure}[t]
  \begin{center}
  \includegraphics[width=0.95\linewidth]{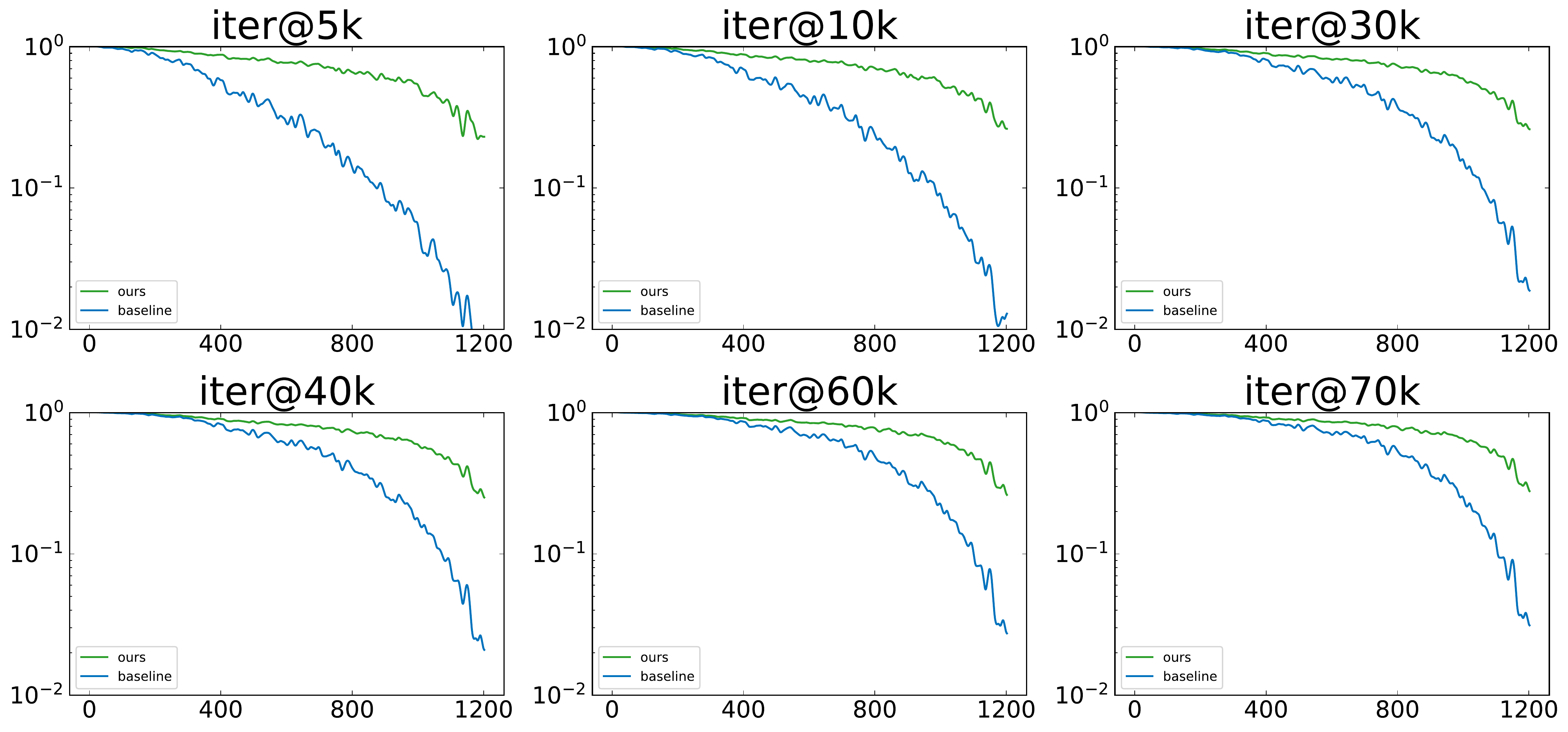}
  \end{center} 
     \caption{The accumulated gradients ratio of positives to negatives. Models are trained with a total of 75k iterations. We show the values at different training iterations. We compare the accumulated gradients of two models, Mask R-CNN with Sigmoid loss and Sigmoid-EQL.}
  \label{fig:vis_balance_ratio}
  \end{figure}

\vspace{0.2cm}
\noindent \textbf{Do we have a more balanced gradient ratio?}  We visualize the gradient ratio of our method Sigmoid-EQL and BCE baseline model during the training process, see Figure \ref{fig:vis_balance_ratio}. The baseline model does not have a balanced ratio for all categories. The positive gradients are overwhelmed by the negative gradients, especially for tail classes,  which makes it hard to detect them. And training longer does not help a lot. In contrast, Sigmoid-EQL preserves a more balanced gradient ratio in the entire training phase.

\vspace{0.2cm}
\label{sec:better_repr}
 \noindent \textbf{Do we have a better representation?} To evaluate the quality of representations trained with our method. We adopt models trained with the Sigmoid-EQL and the CE loss as the pre-trained models. Then we follow the classic decoupled training recipe to re-train the classifier with frozen representations. The results are shown in Table \ref{tab:better_repr}. There are two main observations: Firstly, models initialized with Sigmoid-EQL always achieve a higher AP, resulting in 22.4\% AP \vs 22.0\% AP for cRT, 23.1\% AP \vs 17.0\% AP for LWS, 24.0\% AP \vs 23.1\% AP for BAGS. It verifies that we obtain a better representation by adopting Sigmoid-EQL compared to standard training. This result doubts the claim~\cite{crt2019kang} that re-weighing will hurt the representation. Secondly, the models get marginal improvements or even worse performance after decoupled training. Compared with the end-to-end Sigmoid-EQL (Table \ref{tab:compare_e2e_decouple}) with 23.7\% AP, the AP only increase 0.3 \% after using re-training on BAGS, and the AP drops 1.3 \% and 0.6 \% after the re-training on cRT and LWS respectively. It shows that decoupled training is not always necessary, we can train models with both a balanced classifier and better representations in an end-to-end manner.

\begin{table}[t]
    \centering
    \setlength\tabcolsep{7pt}
    \begin{tabular}{c | c | c c c c c }
          \toprule
          method & EQL & AP & AP\textsubscript{\textit{r}} & AP\textsubscript{\textit{c}} & AP\textsubscript{\textit{f}} & AP\textsubscript{\textit{b}}\\
          \midrule
          cRT~\cite{crt2019kang} & & 22.0 & \textbf{13.5} & 20.8 & 27.1 & 22.1 \\
          cRT~\cite{crt2019kang} & \checkmark & \textbf{22.4} & 13.3 & \textbf{21.7} & \textbf{27.3} & \textbf{22.4} \\
          \midrule
          LWS~\cite{crt2019kang} & & 17.0 & 2.0 & 13.5 & 27.4 & 17.5 \\
          LWS~\cite{crt2019kang} & \checkmark & \textbf{23.1} & \textbf{13.8} & \textbf{22.2} & \textbf{28.1} & \textbf{23.2} \\
          \midrule
          BAGS~\cite{bags2020li} & & 23.1 & 13.1 & 22.5 & 28.2 & 23.7 \\
          BAGS~\cite{bags2020li} & \checkmark & \textbf{24.0} & \textbf{14.6} & \textbf{23.8} & \textbf{28.5} & \textbf{24.5} \\
          \bottomrule
    \end{tabular}
    \caption{Results of various decoupled training methods with different pre-trained models. EQL indicates the pre-trained models are trained with EQL, otherwise with standard training. Only random samplers are used.}
    \label{tab:better_repr}
 \end{table}

 \begin{table}
  \centering
  \small
  \setlength\tabcolsep{6pt}
  \begin{tabular}{l | c |  c c c c c}
         \toprule
         &  AP & AP1 & AP2 & AP3 & AP4 & AP5 \\ 
        \midrule
        Faster-R50 & 43.1 & 26.3 & 42.5 & 45.2 & 48.2 & 52.6 \\
        Grad-Blocking~\cite{eql2020tan} & 45.3 & 32.7 & 44.6 & 47.3 & 48.3 & 53.1 \\
        Sigmoid-EQL & 52.6 & 48.6 & 52.0 & 53.0 & 53.4 & 55.8 \\
        \midrule
        Faster-R101 & 46.0 & 29.2 & 45.5 & 49.3 & 50.9 & 54.7 \\
        Grad-Blocking~\cite{eql2020tan} & 48.0 & 36.1 & 47.2 & 50.5 & 51.0 & 55.0 \\
        Sigmoid-EQL & 55.1 & 51.0 & 55.2 & 56.6 & 55.6 & 57.5 \\
        \bottomrule
  \end{tabular}
  \caption{Results on \textbf{Open Images Challenge 2019} \texttt{val} set. The model Faster R-CNN~\cite{faster_rcnn2015ren} with ResNet-FPN is trained with a schedule of 120k/160k/180k. Categories are grouped into five groups according to the instance number. AP1 is the AP of the first group, where categories have the least annotations, and AP5 is the AP of the last group, where categories have the most annotations. we directly use the hyper-parameters searched in LVIS, }
  \label{tab:openimage}
\end{table}

\vspace{0.2cm}
\label{sec:open_images_exp}
\noindent \textbf{Generalization to other datasets.} 
To verify the generalization ability to other datasets, we conduct experiments on the OpenImages~\cite{openimages2018kuznetsova}. OpenImages is another large-scale object detection dataset with long-tailed distributed categories. We use the data split of challenge 2019, which is a subset of OpenImages V5. The \texttt{train} set consists of 1.7M images of 500 categories. We evaluate our models on the 41k \texttt{val} set. In addition to the standard mAP@IOU=0.5 metric, we also group categories into five groups (100 categories per group) according to their instance numbers and report the mAP within each group respectively. The results are shown in Table~\ref{tab:openimage}. Sigmoid-EQL reaches an AP of 52.6\%, outperforming the baseline model and Grad-Blocking~\cite{eql2020tan} by 9.5\% AP and 7.3\% AP respectively. For the tail group (AP1), the Sigmoid-EQL increases the AP by 22.3 points, which is much more than the improvement of Grad-Blocking (6.4\% AP). Sigmoid-EQL also outperforms Grad-Blocking considerably on the larger ResNet-101 backbone. For both baseline and Grad-Blocking models, there is still a large performance gap between head and tail classes. Sigmoid-EQL brings all categories into a more equal status. It achieves similar accuracy for all category groups. It is worth noting that we tune the hyper-parameter $\lambda$ in Grad-Blocking which puts 250 categories into the tail group for OpenImage. In contrast, the hyper-parameters of Sigmoid-EQL are kept the same as that on LVIS. Those experiments not only show the effectiveness but also the good generalization ability of Sigmoid-EQL.

\subsection{Single-Stage Long-tailed Object Detection}
In this section, we explore how to train a high-performance single-stage object detector successfully in the long-tailed setting. We mainly study the gradient-indicated focal loss, \ie Equalized Focal Loss.

\begin{table*}
  \centering
  \setlength{\tabcolsep}{3.5mm}{
  \begin{tabular}{c|l|ccc|cccc}
    \toprule
    backbone & method & strategy & sampler & epoch & AP & AP$_r$ & AP$_c$ & AP$_f$ \\
    \midrule
    \multirow{11}{*}{ResNet-50} & \textit{two-stage} &  &  &  &  &  &  &  \\
    ~ & Faster R-CNN \cite{faster_rcnn2015ren} & end-to-end & RFS & 24 & 24.1 & 14.7 & 22.2 & 30.5 \\
    ~ & Grad-Blocking \cite{eql2020tan} & end-to-end & RFS & 24 & 25.1 & 15.7 & 24.4 & 30.1 \\
    ~ & Seasaw Loss \cite{seesawloss2021wang} & end-to-end & RFS & 24 & 26.4 & 17.5 & 25.3 & 31.5 \\
    ~ & cRT \cite{crt2019kang} & decoupled & RFS+CBS & 24+12 & 24.8 & 15.9 & 23.6 & 30.1 \\
    ~ & BAGS \cite{bags2020li} & decoupled & RFS+CBS & 24+12 & 26.0 & 17.2 & 24.9 & 31.1 \\
    ~ & NorCal \cite{pan2021model} & post-hoc & RFS & 24 & 26.6 & 18.7 & 25.6 & 31.1 \\
    \cmidrule{2-9}
    ~ & \textit{one-stage} &  &  &  &  &  &  &  \\
    ~ & RetinaNet \cite{focalloss2017lin} & end-to-end & RFS & 24 & 18.5 & 9.6 & 16.1 & 25.0 \\
    ~ & Baseline$^{\dag}$ & end-to-end & RFS & 24 & 25.7 & 14.3 & 23.8 & \textbf{32.7} \\
    ~ & EFL (Ours) & end-to-end & RFS & 24 & \textbf{27.5} & \textbf{20.2} & \textbf{26.1} & 32.4 \\
    \midrule
    \multirow{9}{*}{ResNet-101} & \textit{two-stage} &  &  &  &  &  &  &  \\
    ~ & Faster R-CNN \cite{faster_rcnn2015ren} & end-to-end & RFS & 24 & 25.7 & 15.1 & 24.1 & 32.0 \\
    ~ & Seasaw Loss \cite{seesawloss2021wang} & end-to-end & RFS & 24 & 27.8 & 18.7 & 27.0 & 32.8 \\
    ~ & BAGS \cite{bags2020li} & decoupled & RFS+CBS & 24+12 & 27.6 & 18.7 & 26.5 & 32.6 \\
    ~ & NorCal \cite{pan2021model} & post-hoc & RFS & 24 & 27.8 & 19.4 & 26.9 & 32.5 \\
    \cmidrule{2-9}
    ~ & \textit{one-stage} &  &  &  &  &  &  &  \\
    ~ & RetinaNet \cite{focalloss2017lin} & end-to-end & RFS & 24 & 19.6 & 10.1 & 17.3 & 26.2 \\
    ~ & Baseline$^{\dag}$ & end-to-end & RFS & 24 & 27.0 & 14.4 & 25.7 & \textbf{34.0} \\
    ~ & EFL (Ours) & end-to-end & RFS & 24 & \textbf{29.2} & \textbf{23.5} & \textbf{27.4} & 33.8 \\
    \bottomrule
  \end{tabular}}
  \caption{Main results of EFL compared with other methods on LVIS v1 \texttt{val} split.
           Baseline$^{\dag}$ indicates the improved baseline.
           RFS and CBS indicate the repeat factor sampler and the class balanced sampler, respectively.
           All end-to-end methods are trained by a schedule of 2x with the RFS while the decoupled methods have an additional 1x schedule with the CBS during the fine-tuning stage.}
  \label{tab:main_results}
\end{table*}

\subsubsection{Datasets and Evaluation Metric}
We perform experiments on the LVIS v1.0 dataset. We report box AP instead of mask AP because most single-stage detectors do not have a mask head. Since most experimental results of two-stage methods are based on the Mask R-CNN \cite{maskrcnn2017he} framework, their released detection performance AP$_b$ is affected by the segmentation performance. We report the detection results of those works by rerunning their code with the Faster R-CNN \cite{faster_rcnn2015ren} framework for a fair comparison.

\subsubsection{Implementation Details}
Most training settings are the same as that in two-stage object detection experiments. As one-stage detectors often predict boxes with low scores, we do not filter out any predicted box before NMS (set the minimum score threshold to 0).
The top 300 confident boxes per image are selected as the final detection results. All models are trained with the repeat factor sampler (RFS) by a 2x schedule. For our proposed EFL, we set the balanced factor $\alpha_{\mathrm{t}} = 0.25$ and the base focusing parameter $\gamma_b = 2.0$ which are same as that in focal loss \cite{focalloss2017lin}.
The scaling hyper-parameter $s$ is set to 8.

 \begin{figure*}[ht]
  \begin{center}
  \includegraphics[width=0.95\linewidth]{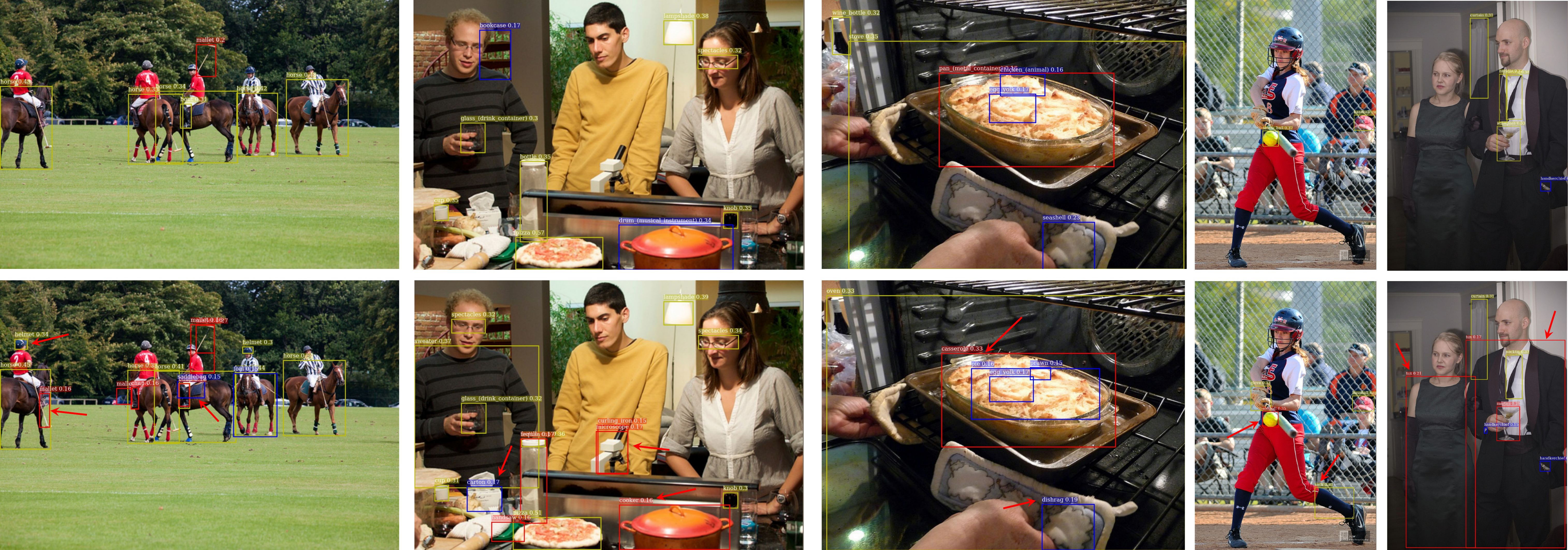}
  \end{center} 
     \caption{\textbf{Equalized Focal Loss} results on LVIS \texttt{val} set. The top row and bottom row are for focal loss and EFL respectively. Red, blue and yellow boxes denote detected rare, common, and frequent categories respectively.}
    \label{fig:vis_od}
\end{figure*}

\vspace{0.2cm}
\noindent
\textbf{Stabilized Setting and Improved Baseline.} We found that there is a large performance gap between the two-stage baseline Faster R-CNN \cite{faster_rcnn2015ren} and the widely-used one-stage detector RetinaNet \cite{focalloss2017lin}.
To bridge this gap, we investigate plenty of one-stage frameworks to establish an improved baseline that is more appropriate for the long-tailed task.
ATSS \cite{atss2020zhang} stands out among those methods with its simplicity and high performance.
From experiments, we discover that the training processes of most one-stage detectors are pretty unstable with result fluctuations and sometimes encountering NaN problems.
Intuitively, the primary culprit is the abnormal gradients in the early training stage caused by severe imbalance problems.
To stabilize the training processes, we adopt the stabilized settings that extend the warm-up iterations from 1000 to 6000 and utilize the gradient clipping with a maximum normalized value of 35.
Meanwhile, in ATSS, we adopt an IoU branch to replace the centerness branch and set the anchor scale from \{8\} to \{6, 8\} with the hyper-parameter $k=18$ to cover more potential candidates.
The combination of stabilized and improved settings is adopted as our improved baseline (named Baseline$^\dagger$).
Unless otherwise stated, EFL is trained with the Baseline$^\dagger$.

\subsubsection{Main Results}
To show the effectiveness of our proposed method, we compare EFL with other works that report state-of-the-art performance.
As demonstrated in Table \ref{tab:main_results}, with ResNet-50-FPN backbone, our proposed method achieves an overall 27.5\% AP, which improves the proposed Baseline$^\dagger$ by 1.8\% AP, and even achieves 5.9 points improvement on rare categories.
The result shows that EFL could handle the extreme positive-negative imbalance problem of rare categories well.
Compared with other end-to-end methods like Grad-Blocking \cite{eql2020tan} and Seesaw Loss \cite{seesawloss2021wang}, our proposed method outperforms them by 2.4\% AP and 1.1\% AP, respectively.
And compared with decoupled training approaches like cRT \cite{crt2019kang} and BAGS \cite{bags2020li}, our approach surpasses them with an elegant end-to-end training strategy (by 2.7\% AP and 1.5\% AP). Meanwhile, compared with post-hoc calibration method NorCal \cite{pan2021model}, EFL still achieves non-trivial improvement (+0.9\% AP). Besides the high performance, we keep the advantages of one-stage detectors like simplicity, rapidity, and ease of deployment.

With the larger ResNet-101-FPN backbone, our approach still performs well on the Baseline$^\dagger$ (+2.2\% AP).
Meanwhile, our approach maintains stable performance improvements compared with all existing methods, whether they are end-to-end or decoupled.
Without bells and whistles, our method achieves 29.2\% AP that establishes a new state-of-the-art.
Notably, the performance of rare categories on the Baseline$^\dagger$ does not gain too much performance improvement from the larger backbone while EFL does (from 20.2\% AP to 23.5\% AP).
It indicates that our proposed method has a good generalization ability across different backbones. Additionally, we show some qualitative analysis in Fig~\ref{fig:vis_od}. EFL outperforms the focal loss by focusing more on the learning of rare and common categories (\ie the red boxes in the figure).

\begin{table}[t]
  \centering
  \setlength{\tabcolsep}{4mm}{
  \begin{tabular}{cc|cccc}
    \toprule
    WF & FF & AP & $AP_r$ & $AP_c$ & $AP_f$ \\
    \midrule
     &  & 25.7 & 14.3 & 23.8 & \textbf{32.7} \\
    \checkmark &  & 26.1 & 15.6 & 24.5 & 32.6 \\
     & \checkmark & 26.2 & 17.7 & 24.7 & 31.5 \\
    \checkmark & \checkmark & \textbf{27.5} & \textbf{20.2} & \textbf{26.1} & 32.4 \\
    \bottomrule
  \end{tabular}}
  \caption{Ablation study of each component in Equalized Focal Loss.
  WF and FF indicate the weighting factor and the focusing factor, respectively.}
  \label{tab:ablation_study_components}
\end{table}

\begin{table}[t]
  \centering
  \setlength{\tabcolsep}{5mm}{
  \begin{tabular}{c|cccc}
    \toprule
    $s$ & AP & AP$_r$ & AP$_c$ & AP$_f$ \\
    \midrule
    0 & 25.7 & 14.3 & 23.8 & \textbf{32.7} \\
    1 & 26.3 & 16.3 & 24.6 & 32.6 \\
    2 & 26.6 & 17.6 & 24.6 &  \textbf{32.7} \\
    4 & 27.3 & 19.9 & 25.5 & 32.6 \\
    8 & \textbf{27.5} & \textbf{20.2} & \textbf{26.1} & 32.4 \\
    12 & 26.5 & 19.9 & 24.6 & 31.6 \\
    \bottomrule
  \end{tabular}}
  \caption{Ablation study of the hyper-parameter $s$.
           $s=8$ is adopted as the default setting in other experiments.}
  \label{tab:s_effect}
\end{table}

\subsubsection{Analysis}
\label{sec:single_stage_analysis}
\noindent
\textbf{Influence of components in EFL.} There are two components in EFL, which are the focusing factor and the weighting factor.
We demonstrate the effect of each component in Table \ref{tab:ablation_study_components}. Both the weighting factor and the focusing factor play significant roles in EFL.
For the focusing factor, it achieves an improvement from 25.7\% AP to 26.2\% AP.
Meanwhile, it brings a significant gain of rare categories with 3.4\% AP improvement, indicating its effectiveness in alleviating the severe positive-negative imbalance problems.
And for the weighting factor, we investigate its influence by setting the focusing factor always be $\gamma_b$ among all categories in EFL.
Thus the function of the weighting factor could also be regarded as a re-weighting approach combined with focal loss. Without the focusing factor, the weighting factor only outperforms the Baseline$^\dagger$ slightly by 0.4\% AP. With the synergy of these two components, EFL dramatically improves the performance of the Baseline$^\dagger$ from 25.7\% AP to 27.5\% AP.

\begin{table}[t]
  \centering
  \setlength{\tabcolsep}{2mm}{
  \begin{tabular}{l|c|cccc}
    \toprule
    method & loss & AP & $AP_r$ & $AP_c$ & $AP_f$ \\
    \midrule
    \multirow{3}{*}{FCOS$^*$ \cite{fcos2019tian}} & FL & 22.6 & 12.7 & 20.9 & \textbf{28.9} \\
    ~ & FL+EQL & 23.0(+0.4) & 14.1 & 21.3 & 28.7 \\
    ~ & EFL & \textbf{23.4 (+0.8)} & \textbf{14.9} & \textbf{21.9} & 28.7 \\
    \midrule
    \multirow{3}{*}{PAA \cite{paa2020kim}} & FL & 23.7 & 14.2 & 21.6 & 30.2 \\
    ~ & FL+EQL & 24.1 +(0.4) & 16.5 & 22.1 & 29.8 \\
    ~ & EFL & \textbf{25.6 +(1.9)} & \textbf{19.8} & \textbf{23.8} & \textbf{30.2} \\
    \midrule
    \multirow{3}{*}{ATSS \cite{atss2020zhang}} & FL & 24.7 & 13.7 & 23.4 & 31.1 \\
    ~ & FL+EQL & 25.2 (+0.5) & 15.0 & 24.3 & 30.8 \\
    ~ & EFL & \textbf{25.8 (+1.9)} & \textbf{18.1} & \textbf{24.5} & \textbf{30.6} \\
    \midrule
    \multirow{2}{*}{Baseline$^{\dag}$} & FL & 25.7 & 14.3 & 23.8 & \textbf{32.7} \\
    ~ & FL+EQL & 26.8(+0.9) & 17.7 & 25.3 & 32.6 \\
    ~ & EFL & \textbf{27.5 (+1.8)} & \textbf{20.2} & \textbf{26.1} & 32.4 \\
    \bottomrule
  \end{tabular}}
  \caption{Results of EFL combined with other one-stage object detectors.}
  \label{tab:one_stage_with_efl}
\end{table}

\vspace{0.2cm}
\noindent
\textbf{Influence of the Hyper-parameter.} Too many hyper-parameters will affect the generalization ability of a method.
In this paper, our proposed EFL only has one hyper-parameter $s$ which is also one of the advantages of our work.
We study the influence of $s$ with different values and find that $s=8$ achieves the best performance.
As presented in Table \ref{tab:s_effect}, almost all $s \geq 0$ could increases the performance of the Baseline$^\dagger$.
What's more, $s$ keeps relatively high performance in a wide range.
It indicates that our EFL is hyper-parameter insensitive.

\vspace{0.2cm}
\noindent
\textbf{Combined with other one-stage detectors.} To demonstrate the generalization ability of EFL across different one-stage detectors, we combine it with FCOS$^*$~\footnote{FCOS$^{*}$ indicates that the reported FCOS result is trained with a center-sampling strategy \cite{atss2020zhang}}, PAA, ATSS, and our Baseline$^\dagger$, separately. We also use the combination of Sigmoid-EQL and Focal loss, referred to as FL+EQL.
As presented in Table \ref{tab:one_stage_with_efl}, EFL and FL+EQL both performs well with all those one-stage detectors.
We notice that EFL consistently outperforms FL+EQL due to its tight bond with focal loss.
EFL maintains a stable large performance gain on overall AP and larger improvement on rare categories.
Those experiments show EFL's strength in settling the long-tailed distribution problem.
What's more, we also investigate whether our proposed EFL and EQFL could work well with the advanced YOLOX$^{*}$ \cite{ge2021yolox} detectors (more implementation details could be found in our supplementary material). As presented in Table \ref{tab:yolox_results}, combined with the YOLOX$^*$ medium model, EFL and EQFL reach the overall AP of 30.0\% and 31.0\% respectively, outperforming the baseline FL and QFL by a large margin and bringing great improvement on the rare categories (about +16.7\% AP from FL to EFL). The results demonstrate that our proposed method is a very practical approach that could greatly alleviate the long-tailed imbalance problem for almost all one-stage detectors.

\begin{table}[t]
  \centering
  \setlength{\tabcolsep}{2mm}{
  \begin{tabular}{l|c|c|cccc}
    \toprule
    model & loss & YOLOX$^{*}$ & AP & AP$_r$ & AP$_c$ & AP$_f$ \\
    \midrule
    \multirow{5}{*}{small} & Sigmoid &  & 15.2 & 2.9 & 11.6 & 24.7 \\
    \cmidrule{2-7}
    ~ & FL & \checkmark & 18.5 & 3.6 & 15.7 & \textbf{28.2} \\
    ~ & \textbf{EFL(Ours)} & \checkmark & \textbf{23.3} & \textbf{18.1} & \textbf{21.2} & 28.0 \\
    \cmidrule{2-7}
    ~ & QFL & \checkmark & 22.5 & 11.0 & 20.6 & \textbf{29.7} \\
    ~ & \textbf{EQFL(Ours)} & \checkmark & \textbf{24.2} & \textbf{16.3} & \textbf{22.7} & 29.4 \\
    \midrule
    \multirow{5}{*}{medium} & Sigmoid &  & 20.9 & 5.3 & 17.6 & 31.5 \\
    \cmidrule{2-7}
    ~ & FL & \checkmark & 25.0 & 7.1 & 23.5 & 34.4 \\   
    ~ & \textbf{EFL(Ours)} & \checkmark & \textbf{30.0} & \textbf{23.8} & \textbf{28.2} & \textbf{34.7} \\
    \cmidrule{2-7}
    ~ & QFL & \checkmark & 28.9 & 16.8 & 27.2 & 36.1 \\
    ~ & \textbf{EQFL(Ours)} & \checkmark & \textbf{31.0} & \textbf{24.0} & \textbf{29.1} & \textbf{36.2} \\
    \bottomrule
  \end{tabular}}
  \caption{Results of the YOLOX \cite{ge2021yolox} detectors combining with EQL and EQFL on the LVIS v1.
  All experiments are trained from scratch by 300 epochs with RFS. The YOLOX$^*$ indicates the enhanced YOLOX detector with our improved settings (see the supplementary material).}
  \label{tab:yolox_results}
\end{table}

\vspace{0.2cm}
\label{sec:oid_results}
\noindent \textbf{Generalization on OpenImages}. As presented in Table \ref{tab:oid_results}, our proposed improved baseline (Baseline$^\dagger$) greatly improves the performance of one-stage detectors.
It brings 11.2\% AP improvement compared with the widely used RetinaNet.
Combined with the Baseline$^\dagger$, our proposed EFL achieves an overall AP of 51.5\% with the ResNet-50 backbone, which outperforms the two-stage baseline Faster R-CNN and the Baseline$^\dagger$ by 8.4\% AP and 8.2\% AP, respectively.
What's more, EFL significantly improves the performance of the rare categories with an improvement of 33.4\% AP on the AP1 split compared with the Baseline$^\dagger$. 
With the larger ResNet-101 backbone, our proposed method still performs well and brings significant AP gains.
Meanwhile, it maintains an excellent performance on rare categories.
All experimental results demonstrate the strength and generalization ability of our method.

\begin{table}[t]
  \centering
  \setlength{\tabcolsep}{2mm}{
  \begin{tabular}{l|c|ccccc}
    \toprule
    method & AP & AP$1$ & AP$2$ & AP$3$ & AP$4$ & AP$5$ \\
    \midrule
    \textit{R-50 w/ FPN}  &  &  &  &  &  &  \\
    Faster R-CNN \cite{faster_rcnn2015ren} & 43.1 & 26.3 & 42.5 & 45.2 & 48.2 & \textbf{52.6} \\
    RetinaNet \cite{focalloss2017lin} & 32.1 & 21.0 & 34.0 & 35.4 & 35.6 & 34.2 \\
    Baseline$^{\dag}$ \cite{atss2020zhang} & 43.3 & 19.4 & 44.3 & 49.5 & \textbf{50.6} & 52.2 \\
    \textbf{EFL (Ours)} & \textbf{51.5} & \textbf{52.8} & \textbf{52.9} & \textbf{50.8} & 50.2 & 50.9 \\
    \midrule
    \textit{R-101 w/ FPN}  &  &  &  &  &  &  \\
    Faster R-CNN \cite{faster_rcnn2015ren} & 46.0 & 29.2 & 45.5 & 49.3 & 50.9 & \textbf{54.7} \\
    RetinaNet \cite{focalloss2017lin} & 35.8 & 26.4 & 38.9 & 38.3 & 38.1 & 36.8 \\
    Baseline$^{\dag}$ \cite{atss2020zhang} & 44.7 & 19.6 & 46.6 & 51.1 & \textbf{52.3} & 53.4 \\
    \textbf{EFL (Ours)} & \textbf{52.6} & \textbf{53.4} & \textbf{53.8} & \textbf{51.4} & 51.8 & 52.3 \\
    \bottomrule
  \end{tabular}}
  \caption{Results on OpenImages Challenge 2019 \texttt{val} split.}
  \label{tab:oid_results}
\end{table}

\begin{table}[t]
  \centering
  \setlength{\tabcolsep}{2.6mm}{
  \begin{tabular}{l|ccc|ccc}
    \toprule
    loss & AP & AP$_{50}$ & AP$_{75}$ & AP$_s$ & AP$_m$ & AP$_l$ \\
    \midrule
    FL & 42.3 & 61.0 & 45.7 & 26.7 & 46.2 & 53.0  \\
    EFL(s=2) & 42.4 & 61.0 & 46.1 & 26.5 & 46.2 & 53.2 \\
    EFL(s=4) & 42.3 & 61.2 & 45.6 & 26.6 & 46.1 & 52.9 \\
    EFL(s=8) & 42.2 & 60.8 & 45.6 & 26.4 & 46.1 & 52.7 \\
    \bottomrule
  \end{tabular}}
  \caption{Results in the COCO dataset.
  All results are from the improved baseline with the ResNet-50 backbone.
  The models are trained by a 2x schedule with the random sampler.}
  \label{tab:coco_results}
\end{table}

\vspace{0.2cm}
\noindent \textbf{Performance on Balanced Dataset}. As we claimed, EFL is equivalent to the focal loss in the balanced data scenario. To verify this analysis, we conduct experiments on the MS-COCO dataset with balanced data distribution.
As presented in Table \ref{tab:coco_results}, the scaling factor $s$ has little effect in the COCO dataset, and all results with EFL achieve comparable performance with the focal loss.
This indicates that our proposed EFL could maintain good performance under the balanced data distribution. This distribution-agnostic property enables EFL to work well with real-world applications in different data distributions.


\begin{table*}[t]
  \centering
  \setlength{\tabcolsep}{3.5mm}
  \begin{tabular}{l|c c c c | c c c c}
    \toprule
     \multirow{2}[2]{*}{method} & \multicolumn{4}{c|}{ResNet-50} & \multicolumn{4}{c}{ResNeXt-50} \\
    \cmidrule(l){2-9}
     & Overall & Many & Medium & Few & Overall & Many & Medium & Few \\
    \midrule
    CE & 41.8 & 64.4 & 33.8 & 5.9 & 44.5 & 65.9 & 37.7 & 8.2 \\
    BCE & 39.6 & 63.7 & 30.4 & 4.0 & 42.7 & 65.9 & 34.4 & 6.4\\
    Logit Adjustment~\cite{logit_adjustment2021Aditya} & 47.3 & 59.2 & 44.7 & 23.0 & 49.6 & 60.9 & 47.3 & 26.1 \\
    Logit Adjustment (post-hoc)~\cite{logit_adjustment2021Aditya} & 47.6 & 59.2 & 45.3 & 22.9  & 49.9 & 61.4 & 47.5 & 26.3 \\
    Balanced Softmax~\cite{BalMS2020ren} & 47.6 & 59.4 & 45.1 & 22.8 & 50.0 & 61.4 & 47.6 & 26.2 \\
    LADE & 47.8 & 59.5 & 45.4 & 23.4 & 50.3 & 61.5 & 47.8 & 27.1 \\
    cRT~\cite{crt2019kang} & 47.6 & 58.6 & 44.7 & 26.8 & 49.4 & 61.4 & 46.1 & 27.0 \\
    LWS~\cite{crt2019kang} & 47.8 & 57.6 & 45.1 & 29.5 & 50.2 & 60.3 & 47.4 & 31.2 \\
    DisAlign~\cite{distalign2021zhang} &47.7 & 55.5 & 46.0 & 31.6 & 50.2 & 58.8 & 48.1 & 33.4 \\
    DIVE~\cite{dive2021he} & 49.9 & 62.4 & 47.4 & 23.2 & 52.7 & 64.4 & \textbf{50.5} & 27.4 \\
    Seesaw Loss~\cite{seesawloss2021wang} & 49.3 & 66.2 & 44.1 & 19.4 & 50.9 & 67.1 & 46.0 & 22.0 \\
    \midrule
    Sigmoid-EQL & 45.6 & \textbf{66.6} & 38.4 & 11.4 & 47.4 & \textbf{67.5} & 40.8 & 13.4\\
    Softmax-EQL & \textbf{51.3} & 58.9 & \textbf{49.1} & \textbf{39.2} & \textbf{53.0} & 61.7 & 49.8 & \textbf{39.5} \\
    \bottomrule
  \end{tabular}
  \caption{Results on ImageNet-LT with ResNet-50 and ResNeXt-50 as the bockbones}
  \label{tab:imagenet_lt}
\end{table*}

\subsection{Long-tailed Image Classification.}
To demonstrate the generalization ability of the equalization losses when transferring to other tasks, we also evaluate our method on long-tailed image classification datasets.
Different from object detection, the image classification task requires a single predicted category as output, in which case we should take the cross-category rank into account. We found that the sigmoid loss is not suitable for this task, so we use the Softmax-EQL in image classification.

\subsubsection{Datasets and Evaluation Metric}
We conduct experiments on three major benchmarks to evaluate the effectiveness of our proposed method.

\textbf{ImageNet-LT} \cite{oltr2019liu} is a long-tailed version of ImageNet-2012 \cite{imagenet2009deng}, which contains 1000 categories with images number ranging from 1280 to 5 images for each category. There are 116k images for training and 50k images for testing. 

\textbf{iNaturalist2018} \cite{van2018inaturalist} is a real-world large-vocabulary dataset. There are 437.5K images from 8,142 categories, which suffer from severe long-tailed imbalances.

\textbf{Place-LT} \cite{oltr2019liu} is also a commonly used benchmark for long-tailed recognition. It is a subset artificially truncated from the Places2 \cite{zhou2017places}. Its number of images per class ranges from 4980 to 5 among 365 categories.

For all datasets, besides the metric of total average accuracy, we present the average accuracy of many-shot ($\geq $100 images), medium-shot (20$\sim$100 images), and few-shot ($\leq$ 20 images) splits to better understand the improvement of tail classes.

\subsubsection{Implementation Details}
We implement our proposed equalization losses based on the settings of Seesaw Loss \cite{seesawloss2021wang} and cRT \cite{crt2019kang}. We notice that some methods have their own codebase (\eg \cite{distalign2021zhang}) and training setting (\eg \cite{lade2021hong, logit_adjustment2021Aditya}). For a fair comparison, we re-implement all compared methods following the common practice settings of cRT \cite{crt2019kang}. In specific, we use the SGD optimizer with momentum 0.9, batch size 512 (256 for Place-LT), and weight decay 5e-4 (1e-4 for iNaturalist2018). We adopt the cosine learning schedule where the learning rate gradually decays from 0.2 to 0. Random resized crop, horizontal flip, and color jitter (iNaturalist2018 except for this item) are used as data augmentation. We report 90 epochs experimental results on ImageNet-LT (ResNet-50 and ResNeXt-50~\cite{resnext2017xie}) and iNaturalist2018 (ResNet-50). For Place-LT, we choose ResNet-152 (pre-trained on the full ImageNet2012) as the backbone network. We train the Place-LT for 30 epochs with everything frozen except the last layer and classifier, which are all aligned with \cite{crt2019kang}.

\begin{table}[t]
  \centering
  \setlength{\tabcolsep}{2.5mm}
  \begin{tabular}{l|c c c c }
    \toprule
    method & Overall & Many & Medium & Few \\
    \midrule
    CE & 31.5 & 46.6 & 28.3 & 11.0 \\
    BCE & 30.2 & 47.7 & 25.7 & 8.5 \\
    Logit Adj~\cite{logit_adjustment2021Aditya} & 39.7 & 42.6 & 40.2 & 33.0\\
    Logit Adj(post)~\cite{logit_adjustment2021Aditya} & 39.3 & 43.0 & 40.1 & 30.6   \\
    Balanced Softmax~\cite{BalMS2020ren} & 39.6 & 42.6 & 40.1 & 32.9  \\
    LADE~\cite{lade2021hong} & 39.6 & 42.7 & 40.0 & 33.0 \\
    cRT~\cite{crt2019kang} & 37.9 & 44.2 & 38.2 & 25.4  \\
    LWS~\cite{crt2019kang} & 39.6 & 42.2 & 40.3 & 33.0 \\
    DisAlign~\cite{distalign2021zhang} & 39.6 & 42.2 & 41.3 & 33.0 \\
    DIVE~\cite{dive2021he} & 39.8 & 42.8 & 40.3 & 33.1 \\
    Seesaw Loss~\cite{seesawloss2021wang} & 37.0 & \textbf{47.9} & 34.7 & 22.4 \\
    \midrule
    Sigmoid-EQL & 34.2 & 45.0 & 29.9 & 24.1 \\
    Softmax-EQL & \textbf{40.8} & 39.1 & \textbf{41.5} & \textbf{42.6}\\
    \bottomrule
  \end{tabular}
  \caption{Results on Places-LT with pre-trained ResNet152}
  \label{tab:place_Lt}
\end{table}

\subsubsection{Main Results}
Firstly, we present the ImageNet-LT results in Table \ref{tab:imagenet_lt}. As the results show, our proposed Softmax-EQL achieves the highest accuracy on different backbones with significant improvements on the tail categories, outperforming current state-of-the-art methods. In particular, Softmax-EQL surpasses the performance of the number-based adjustment methods (\eg Seesaw Loss \cite{seesawloss2021wang}, LADE \cite{lade2021hong}, Logit Adjustment \cite{logit_adjustment2021Aditya}, and Balanced Softmax\cite{BalMS2020ren}), proving that the gradient-based adjustment method is a reliable choice for long-tailed recognition. Meanwhile, Softmax-EQL also outperforms other multi-training-stage methods (\eg cRT \cite{crt2019kang}, LWS\cite{crt2019kang}, and DisAlign\cite{distalign2021zhang} are finetuned from the CE pre-trained model, while DIVE\cite{dive2021he} mimics the features from the Balanced Softmax). Meanwhile, Sigmoid-EQL outperforms the BCE baseline by a significant margin, especially on tail categories. However, we can see that the sigmoid-based losses are not comparable with softmax-based losses. This is because, for each instance, the sigmoid-based losses do not take the cross-category competition into account. Thus they are not a good choice for long-tailed classification tasks. 

\begin{table}[t]
  \centering
  \setlength{\tabcolsep}{2.5mm}
  \begin{tabular}{l|c c c c }
    \toprule
    method & Overall & Many & Medium & Few \\
    \midrule
    CE & 61.7 & 72.9 & 62.8 & 57.4 \\
    BCE & 62.1 & \textbf{73.3} & 63.1 & 58.0  \\
    Logit Adj~\cite{logit_adjustment2021Aditya} & 66.0 & 64.8 & 65.7 & 66.7 \\
    Logit Adj(post)~\cite{logit_adjustment2021Aditya} & 66.3 & 65.1 & 66.2 & 66.7 \\
    Balanced Softmax~\cite{BalMS2020ren} & 66.0 & 64.7 & 66.0 & 66.5 \\
    LADE~\cite{lade2021hong} & 66.3 & 65.4 & 66.1 & 66.9 \\
    cRT~\cite{crt2019kang} & 64.8 & 69.4 & 65.5 & 62.7 \\
    LWS~\cite{crt2019kang} & 65.9 & 63.7 & 66.2 & 66.0 \\
    DisAlign~\cite{distalign2021zhang} & 65.9 & 63.7 & 66.2 & 66.0 \\
    DIVE~\cite{dive2021he} & 69.1 & 70.3 & 70.0 & 67.8 \\
    Seesaw Loss~\cite{seesawloss2021wang} & 65.0 & 73.0 & 66.4 & 61.3 \\
    \midrule
    Sigmoid-EQL & 62.9 & 73.2 & 63.6 & 59.3 \\
    Softmax-EQL & 66.8 & 63.2 & 66.3 & 68.4 \\
    DIVE-Softmax-EQL & \textbf{70.4} & 68.7 & \textbf{70.4} & \textbf{70.8}\\
    \bottomrule
  \end{tabular}
  \caption{Results on iNaturalist2018 with ResNet-50}
  \label{tab:inat}
\end{table}

 We also conduct experiments on Place-LT and iNaturalist2018. The results are demonstrated in Table \ref{tab:place_Lt} and Table \ref{tab:inat}, respectively. Softmax-EQL still achieves impressive performance on these datasets, demonstrating the generalization ability of our equalization losses on different recognition benchmarks. Furthermore, we propose to mimic the Balanced Softmax features using the Softmax-EQL (called DIVE-Softmax-EQL) following the DIVE strategy. As presented in Table \ref{tab:inat}, DIVE-Softmax-EQL outperforms DIVE by 1.3 points of accuracy, with an improvement of 3.0 points on tail categories. 


\begin{table*}[ht]
  \centering
  \small
  \setlength{\tabcolsep}{2mm}{
  \begin{tabular}{l |c|c|c c c c | c c c c}
    \toprule
    model & backbone & loss &mIoU & mIoU$_1$ & mIoU$_2$ & mIoU$_3$ & mAcc & mAcc$_1$ & mAcc$_2$ & mAcc$_3$ \\
     \midrule
     \multirow{8}{*}{PSPNet~\cite{pspnet2017zhao}} & \multirow{4}{*}{R50} & CE & 31.11 & 47.86 & 29.83 & 15.59 & 39.72 & 61.30 & 39.15 & 18.71 \\
     &  & Softmax-EQL & 33.01 &  48.32 & 30.79 & 19.92 & 46.10 & 64.44 & 46.79 &27.07 \\
     & & BCE & 30.78 & 48.02 & 29.22 & 15.11 & 39.07 & 61.36 & 38.32 & 17.53 \\
     &  & Sigmoid-EQL & 32.87 &  48.68 & 30.52 & 19.40 & 41.56 & 61.70 & 39.95 & 23.01 \\
     \cmidrule(l){2-11}
     & \multirow{4}{*}{R101} & CE & 34.39 & 50.49 & 33.32 & 19.34 & 43.44 & 64.45 & 42.09 & 23.77 \\
     &  & Softmax-EQL & 35.65 & 50.49 & 34.67 & 21.77 & 48.67 & 66.37 & 49.89 & 29.73 \\
     & & BCE & 33.27 & 50.08 & 32.26 & 17.47 & 41.91 & 63.48 & 41.18 & 21.06 \\
     &  & Sigmoid-EQL & 35.35 & 50.68 & 33.66 & 21.70 & 44.73 & 64.08 & 43.19 & 26.94 \\
     \midrule
     \multirow{8}{*}{DeeplabV3+~\cite{deeplabv3plus2018chen}} & \multirow{4}{*}{R50} & CE & 32.06 & 49.22 & 31.96 & 14.99 & 40.93 & 63.57 & 41.84 & 17.4 \\
     &  & Softmax-EQL & 33.16 & 49.12 & 32.22 & 18.13 & 45.13 & 64.89 & 46.93 & 23.56 \\
     & & BCE & 31.58 & 48.69 & 32.36 & 13.69 & 40.27 & 62.86 & 41.93 & 16.02 \\
     &  & Sigmoid-EQL & 32.84 & 49.21 & 32.26 & 17.06 & 42.86 & 63.39 & 43.78 & 21.43 \\
     \cmidrule(l){2-11}
     & \multirow{4}{*}{R101} & CE & 34.69 & 50.92 & 34.34 & 18.80 & 44.09 & 65.10 & 44.85 & 22.33 \\
     &  & Softmax-EQL & 35.08 & 50.98 & 34.59 & 19.67 & 47.42 & 66.81 & 49.15 & 26.32 \\
     & & BCE & 33.43 & 50.85 & 33.92 & 15.52 & 42.57 & 64.72 & 44.15 & 18.84 \\
     &  & Sigmoid-EQL & 36.12 & 51.72 & 35.60 & 21.05 & 46.48 & 66.08 & 47.07 & 26.29 \\
    
    \midrule
  \end{tabular}}
  \caption{Results of \textbf{ADE20K-LT}. We also report grouped mIoU and grouped mACC, where the last value is for tail categories and the first value for head categories.}
  \label{tab:ade20k_lt_results}
\end{table*}

\begin{table*}[ht]
  \centering
  \small
  \setlength{\tabcolsep}{2mm}{
  \begin{tabular}{l |c|c|c c c c | c c c c}
    \toprule
    model & backbone & loss &mIoU & mIoU$_1$ & mIoU$_2$ & mIoU$_3$ & mAcc & mAcc$_1$ & mAcc$_2$ & mAcc$_3$ \\
     \midrule
     \multirow{8}{*}{PSPNet~\cite{pspnet2017zhao}} & \multirow{4}{*}{R50} & CE & 41.52 & 52.67&38.61&33.28 &  52.07 & 65.57&48.66&41.80 \\
     &  & Softmax-EQL & 42.11 & 53.27&39.20&33.84 & 58.67 &  68.80&55.77&51.44 \\
     &  & BCE & 41.21 & 53.00&37.65&32.99 & 51.44 & 66.06&46.94&41.31 \\
     &  & Sigmoid-EQL & 41.95 & 52.74&38.69&34.42 & 53.32 & 66.23&48.97&44.75 \\
     
     \cmidrule(l){2-11}
     & \multirow{4}{*}{R101} & CE &  43.99 & 54.60&41.19&36.18 & 55.45 & 67.86&52.49&46.00  \\
     &  & Softmax-EQL & 44.13 & 54.50&41.40&36.49 & 60.83 & 69.50&58.63&54.36  \\
     &  & BCE & 43.39 & 54.60&40.33&35.25 & 54.19 & 67.60&50.44&44.55 \\
     &  & Sigmoid-EQL &  44.37 & 54.68&41.80&36.64 & 55.66 & 67.74&52.21&47.04 \\
     
     \midrule
     \multirow{8}{*}{DeeplabV3+~\cite{deeplabv3plus2018chen}} & \multirow{4}{*}{R50} & CE &  43.34 & 54.09&40.52&35.40 & 55.11 & 67.83&52.30&45.20\\
     &  & Softmax-EQL &  43.67 & 53.79&40.80&36.40 & 59.44 & 69.47&57.04&51.80 \\
     &  & BCE  & 43.09 & 54.42&40.67&34.17 & 54.85 & 68.27&52.33&43.94 \\
     &  & Sigmoid-EQL & 43.74 & 54.57&41.12&35.53 & 55.38 & 67.67&52.72&45.77 \\
     
     \cmidrule(l){2-11}
     & \multirow{4}{*}{R101} & CE & 44.62 & 55.48&42.60&35.77 & 56.34 & 68.92&54.28&45.83 \\
     &  & Softmax-EQL & 44.49 & 55.54&42.80&35.12 & 60.51 & 70.43&60.25&50.85 \\
     &  & BCE & 44.40 & 55.46&42.65&35.10 & 56.17 &  68.99&54.13&45.39\\
     &  & Sigmoid-EQL & 45.96 & 55.87&43.56&38.44 & 58.44 &  69.51&56.23&49.59 \\
    \midrule
  \end{tabular}}
  \caption{Results on \textbf{ADE20K}.}
  \label{tab:ade20k_results}
\end{table*}

\subsection{Long-tailed Semantic Segmentation.}
Semantic segmentation is another important visual recognition task in computer vision. Semantic segmentation is doing a dense per-pixel classification, which also involves cross-category competition. There is a long-tailed problem in semantic segmentation since the head categories often occupy much more pixels. To show the versatility of equalization losses, we conduct experiments on large-scale semantic segmentation datasets with various models.

\subsubsection{Datasets and Evaluation Metric}
ADE20K~\cite{ade20k2017zhou} is a large scale dataset for semantic segmentation. It has 20,196 training images and 2k testing images. There is a total of 150 categories whose annotations are not carefully picked. So there naturally exists an imbalanced pixel ratio between categories.

Although the pixel ratio is imbalanced between categories, the least category still has annotations in over 400 images. To simulate a more severe long-tailed distribution. We sample the original ADE20K dataset to construct the ADE20K-LT dataset. Specifically, we first calculate the sample ratio for each category with a exponential distribution: $e^{-0.01i}$, $i \in [0,150)$ is the category index. For instance, the last category, index 149, has the sample ratio of 22\%; the first category, index 0, has the sample ratio of 0\%; We start the sampling process from the last category to the first category. With the calculated sample ratio of a given category $j$, we randomly sample images from all images that contain category $j$. More details about the dataset construction process could be found in the appendix. Finally, we generate the long-tailed version of ADE20K, namely ADE20K-LT. It contains 11,578 training images and the same 2k test images as the original ADE20K. Besides the metric mIoU and mAcc, we evenly divide 150 categories into three groups and report the mIoU and mAcc of each group.

\subsubsection{Implementation Details}
We conduct experiments on two strong models: PSPNet~\cite{pspnet2017zhao} and DeeplabV3+~\cite{deeplabv3plus2018chen}. We use a crop size of 512$\times$512 in training. Synchronized Batch Normalization is adopted in the ResNet backbone. Models are trained with a total of 80k iterations for ADE20K (or 40k iterations for ADE20k-LT) using the poly learning rate scheduler. The total batch size is 16 and the initial learning rate is 0.01.

\subsubsection{Main Results}
First of all, we present the result of ADE20K-LT in Table~\ref{tab:ade20k_lt_results}. We choose two popular and powerful models, PSPNet~\cite{pspnet2017zhao} and DeeplabV3+~\cite{deeplabv3plus2018chen}. We choose the CE and BCE loss as baseline loss functions. The CE loss achieves higher results than the BCE loss. This is because that semantic segmentation needs a single predicted category per pixel, and the CE loss aligns with the requirement better. Replacing with Sigmoid-EQL or Softmax-EQL both improves the performance significantly. The Softmax-EQL outperforms the baseline CE loss by a large margin in mAcc. This demonstrates that gradient-based margin calibration can benefit tasks that require inter-category competition. However, the Sigmoid-EQL and Softmax-EQL give similar improvements on mIoU, showing that a higher accuracy does not mean a higher mIoU. An auxiliary loss, \eg IOU-based loss may mitigate this problem.  Next, we report the experimental result of ADE20K in Table~\ref{tab:ade20k_results}. Since ADE20k naturally exists the long-tailed problem, we can see that the later groups have lower mIoU and mAcc (\ie mIoU$_3$ and mAcc$_3$).
The Sigmoid-EQL and Softmax-EQL still outperform the baseline loss functions. For the largest model, DeeplabV3+ with ResNet-101, the Sigmoid-EQL achieve non-trivial improvement, 1.5 points on mIoU and 2.3 points on mAcc, respectively.

\section{Conclusion}
In this work, we systematically study the gradient imbalance problem in long-tailed data distribution. We show that this problem exists consistently in different visual tasks and datasets. We find the gradient statistic is able to serve as a stable and precise indicator for the imbalance status of models. Finally, a new family of gradient-driven loss functions is proposed, namely equalization loss. Extensive experiments demonstrate its effectiveness and generalization ability. We believe that this gradient-driven training idea will serve as an important concept when designing specific algorithms for long-tailed object recognition.


%





\ifCLASSOPTIONcaptionsoff
  \newpage
\fi



%

{
  \bibliographystyle{IEEEtran}
  \bibliography{IEEEabrv,egbib}
} 




%








\end{document}